\documentclass{article}

\usepackage[english]{babel}

\usepackage[letterpaper,top=2cm,bottom=2cm,left=3cm,right=3cm,marginparwidth=1.75cm]{geometry}

\usepackage{setspace}
\usepackage{amsmath}
\usepackage{amssymb}
\usepackage{booktabs}
\usepackage{graphicx}
\usepackage{subcaption}
\usepackage{algorithm}
\usepackage{algpseudocode}
\usepackage[colorlinks=true, allcolors=blue]{hyperref}
\usepackage{xcolor} 
\definecolor{darkgreen}{rgb}{0.0,0.5,0.0}
\definecolor{softBlue}{RGB}{52, 120, 246}
\definecolor{softOrange}{RGB}{255, 159, 67}
\definecolor{softGreen}{RGB}{46, 204, 113}
\usepackage{mhchem}
\usepackage{authblk}
\usepackage[numbers,sort&compress]{natbib}
\usepackage[final]{microtype}
\newcommand{\gaussian}{\mathcal{N}(\mathbf{0}, \mathbf{I})}

\newcommand{\X}{\tilde{\mathbf{X}}}
\newcommand{\Xt}{\tilde{\mathbf{X}}_t}
\newcommand{\Xclean}{\mathbf{X}_0}
\newcommand{\Xprior}{\tilde{\mathbf{X}}_T}

\newcommand{\Xrev}[1]{\tilde{\mathbf{X}}^{(#1)}}     
\newcommand{\sigi}[1]{\sigma^{(#1)}}
\newcommand{\epsi}[1]{\boldsymbol{\epsilon}^{(#1)}}
\newcommand{\sighat}[1]{\hat{\sigma}^{(#1)}}
\newcommand{\Nsteps}{N}
\newcommand{\Ntarget}{N_{\mathrm{target}}}

\newcommand{\Xpred}{\hat{\mathbf{X}}_0}              
\newcommand{\F}{\mathbf{F}}   
\newcommand{\Fpred}{\hat{\mathbf{F}}}                  
\newcommand{\E}{E}                               
\newcommand{\Epred}{\hat{E}}                           
\newcommand{\eps}{\boldsymbol{\epsilon}}         
\newcommand{\epspred}{\hat{\boldsymbol{\epsilon}}}     
\newcommand{\s}{\mathbf{s}}                      
\newcommand{\spred}{\hat{\mathbf{s}}}                  

\title{Generative Pseudo-Force Fields for Molecular Generation}
\author[1,2, *]{Stefaan S. P. Hessmann}
\author[1,2]{Khaled Kahouli}
\author[1,2]{Stefan Gugler}
\author[3,4,1,2]{Michael Plainer}
\author[3,5,6,7]{Frank No\'e}
\author[1,2,8,9]{Klaus-Robert M\"uller}
\author[1,2, *]{Niklas W. A. Gebauer}
\affil[1]{Machine Learning Group, Technische Universit\"at Berlin, Berlin, Germany}
\affil[2]{BIFOLD -- Berlin Institute for the Foundations of Learning and Data, Berlin, Germany}
\affil[3]{Department of Mathematics and Computer Science, Freie Universit\"at Berlin, Berlin, Germany}
\affil[4]{Zuse School ELIZA, Darmstadt, Germany}
\affil[5]{Department of Physics, Freie Universit\"at Berlin, Berlin, Germany}
\affil[6]{Microsoft Research AI4Science, Berlin, Germany}
\affil[7]{Department of Chemistry, Rice University, Houston, USA}
\affil[8]{Max-Planck Institute for Informatics, Saarbr\"ucken, Germany}
\affil[9]{Department of Artificial Intelligence, Korea University, Seoul, South Korea}
\affil[*]{Corresponding authors: \texttt{stefaan.hessmann@tu-berlin.de}, \texttt{niklas.gebauer@tu-berlin.de}}
\begin{document}
\maketitle

\begin{abstract}
Generating stable molecular conformations typically forces a tradeoff between the physical realism of energy-based relaxation and the sampling efficiency of data-driven generative models. While machine learning force fields (MLFFs) can sample stable conformations by relaxing molecular geometries according to physical forces, they require costly \textit{ab-initio} training data. 
Conversely, diffusion models (DMs) learn from equilibrium data alone but are dependent on noise schedules and time-step conditioning. 
In this work, we propose generative pseudo-force fields (GPFFs) to bridge these paradigms by training an MLFF on a quadratic pseudo-potential energy surface relative to reference equilibrium structures.
Because no \textit{ab-initio} calculations are required for the perturbed geometries, non-equilibrium training data can be generated on the fly by perturbing the equilibria with Gaussian noise.
We show that GPFFs constitute a time-step-agnostic variant of variance exploding DMs:
the score comes from the predicted pseudo-forces but because force magnitudes implicitly encode the noise level, no time-step conditioning is needed. 
Our GPFF can hence be used as a drop-in replacement in standard diffusion sampling (ancestral, Heun) but also facilitates more efficient, adaptive variants and an MLFF inspired direct denoising scheme.
Our proposed sampling algorithms support arbitrary structural priors and geometric constraints.
On QM9, GPFF has 100~\% validity at 256 neural function evaluations (NFE) and over 50~\% at just 6 NFE, outperforming diffusion baselines across all samplers.
Combined with custom priors, we showcase the fast and accurate generation process of our method in a molecular editor for a drug design setting, where a molecule is generated in real time.

\end{abstract}

\section{Introduction}
 
\begin{figure}
\centering
\includegraphics[width=1\linewidth]{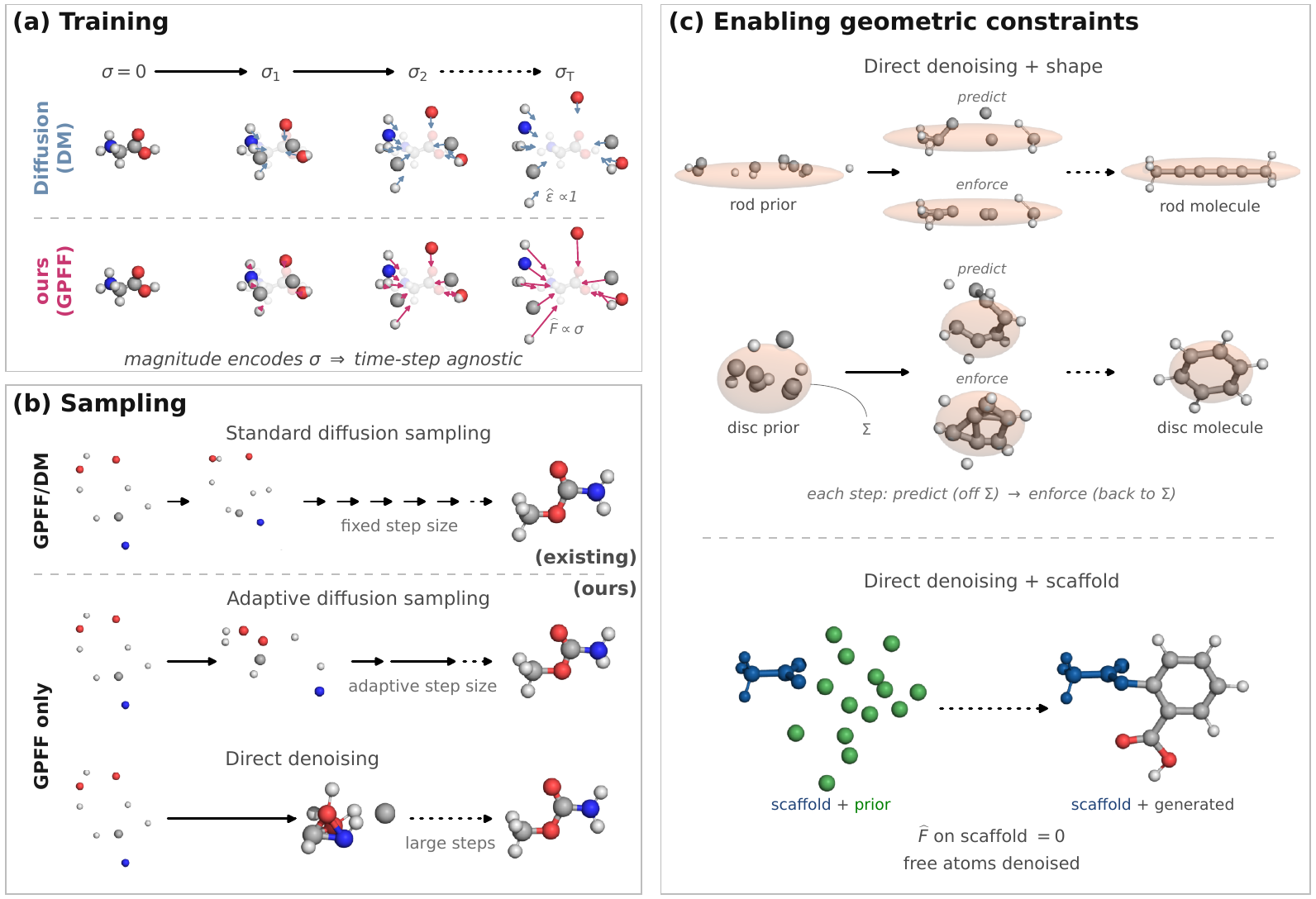}
\caption{\label{fig:overview}
Overview of generative pseudo-force fields (GPFFs).
\textbf{(a) Training.} Both generative models (DM, top) and GPFFs (bottom, ours) are trained on equilibrium structures perturbed with isotropic Gaussian noise of increasing scale $\sigma$.
The DM regresses unit-magnitude noise vectors $\hat{\boldsymbol{\varepsilon}}$ that are invariant to $\sigma$, whereas GPFF learns pseudo-forces $\hat{\mathbf{F}}\propto\sigma$ including the magnitude which implicitly encodes the current noise level, so no time-step conditioning is required.
\textbf{(b) Sampling.} Both families of samplers operate on the same trained GPFF.
\emph{Standard diffusion sampling} is shared with DMs and follows a fixed schedule of equal-sized denoising steps.
\emph{Adaptive diffusion sampling} (ours) reads the noise-level estimate $\sighat{i}$ off the predicted pseudo-force magnitude and rescales each step to correct deviations from the target schedule.
\emph{Direct denoising} (ours) treats sampling as MLFF-style relaxation, taking a few large gradient steps toward the predicted equilibrium.
\textbf{(c) Enabling geometric constraints.} Because GPFF does not require the noise level as an input, intermediate structures can be modified at every step, enabling the extension of direct denoising to arbitrary priors and geometric constraints.
\emph{Shape (top):} starting from a shape-conforming prior, each step first predicts an unconstrained update and then projects the structure back onto the prescribed covariance $\boldsymbol{\Sigma}$, yielding molecules with targeted shape (e.g., rod or disc).
\emph{Scaffold (bottom):} a fixed substructure (blue) is held in place by clamping its pseudo-forces to zero while the remaining free atoms (green) are denoised, enabling scaffold-guided generation.}
\end{figure}
 
The discovery and design of molecules and materials is a fundamental challenge in computational chemistry, with applications ranging from drug design \cite{hajduk2007decade_drug_fragments} to materials science \cite{hautier2011novel_batteries_throughput}.
Beyond the broad exploration of chemical space \cite{rupp2012fast,reymond2015,noe2019boltzmann,von2020exploring,gugler2020}, specific tasks often impose additional constraints, such as targeting desired physical properties \cite{Bhowmik2019inverse_batteries,Freeze2019inverse_catalysts,Gantzer2020inverse_drugs}, incorporating certain substructures \cite{Joshi20213D}, or obeying spatial conditions such as fitting into protein pockets \cite{Prot2010}.
Independent of the target domain, design methods commonly target equilibrium structures, i.e.,\ local minima in the Born--Oppenheimer potential energy surface (PES).

Two approaches dominate the computational discovery of equilibrium structures:
\textit{Geometry relaxation} \cite{jensen2017b,cramer2004,schlegel2011} iteratively follows interatomic forces to a local minimum, demanding multiple computationally expensive calculations \cite{DFT}, which becomes infeasible at scale.
To address these limitations, a plethora of \emph{machine learning force field (MLFF)}~\cite{noe2020machine,unke2021machine_review,UnkeGems,musil2021physics,gugler2025} models, such as kernel methods \cite{rupp2012fast,gdml,sgdml,Chmiela2018Towards,bartok2010gaussian,bartok2017machine} and neural networks \cite{montavon2013machine,schnet,Schuett2018SchNet,spookynet,physnet,painn,keith2021combining,nequip,batatia2025foundation,so3krates,frank2024euclidean,eissler2026simple,frank2026machine} have emerged \cite{bonneau2026breaking}.
Once trained, they enable relaxation from arbitrary inputs with any gradient-descent algorithm.
However, this efficiency relies on extensive training data, with MLFFs requiring datasets of non-equilibrium geometries~\cite{ganscha2025qcml,smith2020ani1x_data,qm7x} labeled with energies and forces from quantum-mechanical calculations. Because these datasets typically only cover a small fraction of the configurational space, MLFFs can fail to reliably relax high-energy inputs, such as randomly sampled or strongly perturbed coordinates \cite{kahouli2024molecular}.

The second approach relies on \emph{generative models}, which bypass force labels entirely by directly learning the distribution of equilibrium structures, enabling substantial progress in computational chemistry \cite{noe2019boltzmann,kohler2020equivariant,garcia2021en_flow,klein2023timewarp,bioemu}.
Auto-regressive models \cite{gschnet,cgschnet} build molecules atom by atom, while diffusion-based denoisers \cite{DDPM_Ho,song2021score} iteratively remove noise from non-equilibrium geometries and currently achieve state-of-the-art results on de novo structure generation \cite{edm,diffbridges,mdm,geoldm,moldiff} and adjacent tasks such as conditional generation \cite{diffbridges}, conformer generation \cite{geodiff,torsional_diff,frank2026sampling}, fragment-based design \cite{hajduk2007decade_drug_fragments,Joshi20213D}, and protein-pocket--targeted generation \cite{Bennett2023,Watson2023}.
However, because diffusion-based models take the current noise level as a network input, sampling is restricted to a fixed schedule and intermediate structures cannot be freely modified between steps.
Both are freedoms that MLFF relaxation naturally provides.
Although various approaches have been developed to accelerate classical diffusion models, e.g., distillation \cite{salimans2022progressive,luhman2021knowledge,frans2025onestep}, better samplers \cite{imp_ddpm,diff_edm}, optimized noise schedules \cite{chen2023importance,lin2024common,kahouli2025}, consistency models \cite{song2023consistency,kim2024consistency,boffi2025buildconsistencymodellearning}, and steering \cite{xie2026enhanced}, the underlying issues remain core challenges in the field.
 
In this work, we propose \textit{generative pseudo-force fields} (GPFFs), a hybrid between MLFFs (inheriting their sampling flexibility) and diffusion models (inheriting their equilibrium-only training).
A GPFF defines a quadratic pseudo-PES whose minima coincide with reference equilibria. Its pseudo-forces and pseudo-energies are cheap to compute on the fly, removing the need for DFT labels.
As in diffusion, training data is generated by perturbing equilibria with Gaussian noise, but training uses the standard MLFF (pseudo-)force-regression loss.
As a consequence, GPFF learns pseudo-forces whose magnitude scales with the noise level, in contrast to diffusion models that learn normalized noise vectors of unit scale at every noise level (Figure~\ref{fig:overview}a).
We show that GPFF is a time-step-agnostic variance-exploding (VE) \cite{song2019score,song2021score} diffusion model: the score is recovered in closed form from the predicted pseudo-forces, and the temperature of the corresponding Boltzmann distribution is set by the diffusion noise level.
 
This dual nature is the central methodological feature of GPFF, and it is reflected directly in the sampling toolbox, for which we propose two complementary families of samplers (Figure~\ref{fig:overview}b):

\textit{From the MLFF perspective}, pseudo-forces drive iterative structure relaxation through direct denoising (DD), a new sampler that iteratively refines towards the equilibrium. 
As in MLFF relaxation, intermediate structures can be modified at every step to enforce constraints, which we illustrate with shape-targeted (DD+shape) and scaffold-based (DD+scaffold) generation (Figure~\ref{fig:overview}c).

\textit{From the diffusion perspective}, the recovered score makes GPFF a drop-in replacement in standard ancestral, Heun, and stochastic-Heun samplers~\cite{song2021score,diff_edm}.
The implicit noise-level encoding additionally yields a cost-free noise-level estimate from the predicted pseudo-forces, which we use to propose adaptive sampling (AS), a novel strategy that corrects for deviations from the true noise schedule and can be applied on top of any of the three standard samplers.
Importantly, all of these samplers operate on the same trained model and can be selected without retraining, depending on the target application.
The MLFF-style DD samplers enable fast relaxation, targeted design, refinement of given structures, and integration into existing MLFF relaxation pipelines.
The diffusion-style samplers cover the standard generative use case, with AS especially beneficial at low step counts.
 
We evaluate GPFF on QM9 \cite{qm9} across all proposed samplers and their non-adaptive baseline counterparts.
With MLFF-inspired DD, GPFF achieves $100\%$ validity at 256 steps and over $50\%$ validity at only 6 steps, outperforming comparable diffusion models.
DD introduces a bias towards compact, spherical molecules in the generated property distributions.
Enforcing geometric shapes on intermediate structures removes this bias without loss of efficiency or validity, and provides an intuitive route to targeted design (rod, disc, sphere shapes).
Furthermore, geometric constraints can be imposed, which we showcase by building a molecular design tool.
With diffusion samplers, GPFF outperforms the diffusion baseline and accurately captures QM9 property distributions, and the adaptive variants further improve both validity and cost, e.g., adaptive Heun improves validity from $87\%$ to $92\%$ at 255 steps.

This work is organized as follows: in Section~\ref{sec:related_work}, we discuss related approaches that connect DMs with MLFFs or remove their noise-level conditioning.
In Section~\ref{sec:methods}, we describe our methodology, starting with the GPFF model and its background followed by the proposed sampling strategies.
We present our results of training a GPFF model on QM9 and evaluating it with all samplers in Section~\ref{sec:results}.
Finally, we provide a conclusion and an outlook on future research directions in Section~\ref{sec:discussion_and_conclusion}.

\section{Related work}
\label{sec:related_work}

Several works have drawn a connection between diffusion-based training and MLFFs.
Zaidi~et~al.\ \cite{zaidi2023pretraining} showed that denoising equilibrium structures perturbed with Gaussian noise is related to learning a force field, and used this objective for property-prediction pretraining.
When training a diffusion model on data that comes from molecular dynamics trajectories, the resulting score function approximates a force field that can drive coarse-grained molecular-dynamics simulation \cite{Arts2023,Plainer2025}.
The closest molecular precedents to GPFF leverage this connection in different ways.
MoreRed \cite{kahouli2024molecular} introduced the term \textit{pseudo-PES} and the corresponding design intuition for molecular relaxation, but trains a noise-conditional diffusion model and recovers the time step with a separate predictor network.
R-DM \cite{woo2026riemannian} trains a noise-agnostic denoiser on slightly perturbed structures for molecular structure optimization with chemical accuracy, using a physics-informed Riemannian manifold of internal coordinates whose metric aligns with molecular energy change.

Furthermore, a growing body of recent work argues that noise-level conditioning is not strictly necessary in denoising-based generative models.
The idea is rooted in blind image denoising \cite{mohan2019robust,kadkhodaie2021stochastic}, where a single denoiser handles a range of noise levels without explicit conditioning.
Sun~et~al.\ \cite{sun2025noise} systematically remove noise conditioning from a wide range of diffusion-based generative models and find graceful degradation, or even improvement, on image benchmarks, while Equilibrium Matching \cite{wang2025equilibrium} and Energy Matching \cite{balcerak2026energy} propose noise-agnostic methods on images that explicitly learn a single, noise-independent gradient field of an energy-like quantity.
Concurrent theoretical work explains why this works: Kadkhodaie~et~al.\ \cite{kadkhodaie2026blind} prove that on low-dimensional data manifolds, a noise-agnostic denoiser can implicitly read the noise level off the noisy input, while Sahraee-Ardakan~et~al.\ \cite{sahraee2026geometry} formalize the resulting generative process as a Riemannian gradient flow on a marginal energy obtained by averaging over noise levels.
GPFF can be viewed as the molecular-domain instance of this noise-agnostic class.
Compared to image-domain members, it is grounded in the MLFF / structure-relaxation analogy with an explicit Cartesian pseudo-PES that yields a closed-form correspondence between the training objective and the VE diffusion perturbation kernel at every noise level.
Compared to R-DM, GPFF works in Cartesian space, supports sampling from arbitrary (including high-noise) initial geometries, and is positioned as a full molecular design tool with flexible geometric constraints rather than a near-equilibrium structure optimizer.

\section{Methods}
\label{sec:methods}
This work encompasses two methodological contributions: generative pseudo-force fields, a novel method for learning machine learning force fields (MLFFs) based on a diffusion-inspired pseudo-energy function, and several sampling strategies for finding equilibrium molecules that leverage the time-agnostic nature of our method compared to conventional diffusion models.
In the following, we separate models and sampling into individual sections.
First, we briefly summarize machine learning force fields and diffusion models (DM) before introducing our generative pseudo-force field (GPFF).
Afterwards, we describe the sampling strategies that we employ for GPFF and the baseline DM.

We denote the Cartesian coordinates of a molecule with $n$ atoms as a matrix $\mathbf{X} \in \mathbb{R}^{n\times3}$.
To keep formulas brief, we use the following conventions.
Any binary operation between matrices and a scalar (e.g., addition or division) is understood entry-wise.
Likewise, the gradient $\nabla_{\mathbf{X}}f(\mathbf{X})$ of a function $f:\mathbb{R}^{n\times3}\rightarrow\mathbb{R}$ is computed entry-wise.
Finally, we write $\mathbf{X}\sim\mathcal{N}(\boldsymbol{\mu}, \sigma^2\mathbf{I})$ for a matrix with entries following an isotropic multivariate Gaussian with mean matrix $\boldsymbol{\mu} \in \mathbb{R}^{n\times3}$ and a variance of $\sigma$ in all $3n$ dimensions.
The corresponding probability density function is

\begin{equation}
    \mathcal{N}(\boldsymbol{\mu}, \sigma^2 \mathbf{I}) =  \frac1Z \exp\left( -\frac{\lVert\mathbf{X}-\boldsymbol{\mu}\rVert_F^2}{2\sigma^2} \right)\,,
\end{equation}

where $Z$ is a normalization constant and $\lVert \cdot \rVert_F$ is the Frobenius norm.

\subsection{Background: Learning potential energy surfaces and probability distributions of equilibrium molecules}

GPFF combines concepts of two existing approaches for finding equilibrium structures.
MLFFs directly approximate the PES $E(\mathbf{X})$ where $E:\mathbb{R}^{n\times3}\rightarrow\mathbb{R}$, while DMs capture a probability distribution over the configurational space $p(\mathbf{X})$ with probability density $p:\mathbb{R}^{n\times3}\rightarrow(0, \infty)$.
In the following, we will briefly summarize these approaches.

\paragraph{Machine learning force fields}

Let $\X \in \mathbb{R}^{n\times3}$ denote the Cartesian coordinates of a non-equilibrium molecular structure.
We can relax $\X$ into an equilibrium configuration by repeatedly doing small gradient descent steps on the PES, which corresponds to following the forces $\F(\X) = -\nabla_{\X}E(\X)$, with recent machine learning approaches aiming to accelerate this by enabling larger steps \cite{bigi2025flashmd,thiemann2025forcefree,ripken2026learning}.
Typically, MLFFs are trained to predict energies $\Epred(\X)$ and forces $\Fpred(\X)$ via the mean squared error loss

\begin{equation}
    \mathcal{L}_{\mathrm{MLFF}} 
    = \mathbb{E}_{\X\sim q_{\mathrm{data}}} \big[ 
     \lambda_{\E}    \|\E(\X) - \Epred(\X) \|^{2}
    +\lambda_{\F}    \|\F(\X) - \Fpred(\X)  \|_F^{2} \big]\,,
    \label{eq:loss_mlff}
\end{equation}

where $\lambda_{\E}$ and $\lambda_{\F}$ are weighting factors for the energy and force components and the expectation over $q_{\mathrm{data}}$ is approximated by an average over the training data samples.
Many MLFF models explicitly compute the forces as the negative gradient of the predicted energy, thereby enforcing energy conservation~\cite{behler2007generalized, schnet, painn, nequip}.

\paragraph{Diffusion models}
Instead of explicitly approximating the PES, generative machine learning models capture the distribution of the training data.
Our GPFF builds on variance exploding DMs~\cite{song2019score,song2021score}, which are probabilistic models that learn to map samples from a simple prior distribution $p_{\mathrm{prior}}=\mathcal{N}\left(\mathbf{0}, \sigma_{T}^2 \mathbf{I}\right)$ to the target distribution $q_{\mathrm{data}}$, e.g.,the manifold of equilibrium molecules.
Given an equilibrium molecule $\Xclean$, we obtain off-manifold samples $\Xt$ by corrupting the Cartesian coordinates with Gaussian noise $\eps_{t} \sim \gaussian$, $\eps_t \in \mathbb{R}^{n \times 3}$:

\begin{equation}
    \Xt = \Xclean + \sigma_t \cdot \eps_{t}\,.
    \label{eq:noising}
\end{equation}

This creates a path of $T$ steps between the data sample and a sample from the prior distribution if we use a monotonically increasing noise schedule $\sigma_1 < \dots < \sigma_T$. Each step $t$ is characterized by a Gaussian \textit{perturbation kernel}, as the closed-form conditional distribution

\begin{equation}
    q(\Xt \mid \Xclean) = \mathcal{N}\left(\Xclean, \sigma_t^2 \mathbf{I}\right) = \frac1Z \exp\left( -\frac{\lVert\Xt-\Xclean\rVert_F^2}{2\sigma_t^2} \right), \;\;t=1,\dots,T \,.
    \label{eq:ve_kernel_one_step}
\end{equation}

The last step approximately gives a sample from the prior $\mathcal{N}\left(\Xclean, \sigma_T^2 \mathbf{I}\right) \approx \mathcal{N}\left(\mathbf{0}, \sigma_{T}^2 \mathbf{I}\right)$ if $\sigma_T \gg \sigma_\mathrm{data}$.
DMs learn to reverse this noising process.
The resulting denoising process involves following the \textit{score} at each step $t$, which is the gradient of the log-probability of the corresponding perturbation kernel

\begin{equation}
    \s(\Xt, t) = \nabla_{\Xt}\log q(\Xt|\Xclean) = \nabla_{\Xt} \left(-\frac{\|\Xt - \Xclean\|_F^2}{2\sigma_t^2} \right) = -\frac{\Xt-\Xclean}{\sigma_t^2} = -\frac{\eps_t}{\sigma_{t}}\,.
    \label{eq:score_ve}
\end{equation}

A common approach is to learn to predict quantities related to the score given the corrupted sample $\Xt$ and either the time step $t$ or the corresponding noise level $\sigma_t$, such as the direction $\eps_{t}$ of the noise that was added to the original sample, the clean data $\Xclean$, or a combination of both to enforce an output of unit variance~\cite{DDPM_Ho,diff_edm,kahouli2025}.
Since the noise level $\sigma_t$ is known, one can reconstruct the score from the predicted noise direction as $\spred(\Xt, t) = -\epspred(\Xt, t)\big/\sigma_t$ \cite{DDPM_Ho, song2021score} or from the predicted clean data using Tweedie's formula $\left(\Xpred(\Xt, t) - \X_t \right) \big/ \sigma_t^2$ \cite{robbins1992empirical, efron2011tweedie, diff_edm}.
We follow the noise parameterization for the baseline DM in our results, using the objective

\begin{equation}
    \mathcal{L}_{\mathrm{DM}} = \mathbb{E}_{t\sim q(t), \Xclean\sim q_{\mathrm{data}}, \Xt\sim q(\Xt|\Xclean)} \left[ \left\| \eps_{t} - \epspred(\Xt, t) \right\|_F^2 \right] \, .
    \label{eq:loss_noise}
\end{equation}

The expectation over $q_{\mathrm{data}}$ is approximated by an average over the training data samples, whereas the remaining expectations are numerically performed by sampling from a distribution over time steps $q(t)$ and applying the perturbation kernel $q(\Xt|\Xclean)$.
The choice of $q(t)$ can have a significant effect on the sampling quality of DMs and we optimize it according to the procedure proposed in EDM~\cite{diff_edm} (see Appendix~\ref{app:noise_levels}).

Once trained, DMs start denoising with a sample from the prior distribution $\Xprior\sim \mathcal{N}\left(\mathbf{0}, \sigma_{T}^2 \mathbf{I}\right)$ used as the endpoint of the forward noising process during training and iteratively follow the score function over a fixed number of steps.
Throughout the trajectory, a noise schedule controls the step sizes and keeps track of the noise level $\sigma_t$ which is needed as an input for model inference and obtaining the score.
Commonly used samplers for denoising include ancestral sampling~\cite{song2021score}, which performs a single score-based update per step, Heun sampling~\cite{diff_edm}, which applies a second-order corrector at the cost of two inference calls per step, and stochastic Heun~\cite{diff_edm}, which additionally injects noise at each step.

\subsection{Generative pseudo-force fields}

Similar to MLFFs, we design GPFF as a PES-based force predictor for molecular relaxation.
However, we replace the DFT-based PES of traditional MLFFs with a pseudo-PES, where energies for non-equilibrium molecules are calculated with respect to the coordinates of a reference equilibrium molecule $\Xclean$:

\begin{equation}
    \E(\X \mid \Xclean) = \|\X - \Xclean\|_{F}^2 \label{eq:pseudo_energy}\,.
\end{equation}

The minimum of this pseudo-PES coincides with a known equilibrium structure on the physical PES and the pseudo-energies must increase with deviation from the reference point. 
As in DMs, we augment equilibrium molecules with Gaussian noise to obtain non-equilibrium structures for training (Eq.~\ref{eq:noising}).
To reduce label ambiguity for molecules with multiple atoms of the same species, we align the atom indexing of the noisy structure to match the reference equilibrium structure (details in Appendix~\ref{app:permutation_alignment})~\cite{klein2023equivariant}.
The pseudo-energy allows us to cheaply compute labels for the obtained pairs without requiring DFT calculations.
The corresponding pseudo-forces are then given as

\begin{equation}
    \F(\Xt \mid \Xclean) = -\nabla_{\Xt} E(\Xt \mid \Xclean)  = -2 (\Xt-\Xclean) = -2\sigma_t\eps_t \, . \label{eq:pseudo_forces}
\end{equation}

The GPFF model learns to predict the pseudo-energies and pseudo-forces given noisy samples $\Xt$ based on the mean squared error objective for MLFFs (Eq.~\ref{eq:loss_mlff}). While for large noise values, i.e., $t \to T$, the resulting noisy structures become physically non-plausible, the pseudo-forces converge to the physical forces as $t \to 0$ \cite{Arts2023,Plainer2025}.
Any existing MLFF model can be used without modifications to the architecture.
In this work, we employ the SchNetPack implementation~\cite{schuett2018schnetpack, schutt2023schnetpack} of the neural network architecture PaiNN~\cite{painn}.
As in conventional MLFF training, pseudo-forces can in principle be computed as derivatives of predicted energies. 
However, we find that this constraint does not lead to improved performance in practice. 
We therefore omit the energy term from the loss (i.e., $\lambda_{E}=0$) and directly learn the pseudo-forces using the gated-equivariant output head of PaiNN.
To prevent that large pseudo-forces dominate the loss, our forces loss weight $\lambda_{\F} = \sigma_t^{-2}$ is reciprocal to the noise level of training samples.
The resulting, simplified loss is

\begin{equation}
    \mathcal{L}_{\mathrm{simple}} = 
        \mathbb{E}_{t\sim q(t),\, \Xclean\sim q_{\mathrm{data}},\, \Xt \sim q(\Xt|\Xclean)} \big[
            \sigma_t^{-2} 
            \|-2\sigma_t\eps_t - \Fpred(\X) \|^2
        \big] \,,
    \label{eq:loss_final}
\end{equation}

where the expectations are computed as in DM training and $q(t)$ is chosen as described in Appendix~\ref{app:noise_levels}.

\paragraph{GPFFs as time-step-agnostic diffusion models}

GPFFs and diffusion models are closely related.
Given a PES $E(\Xt|\Xclean)$, we can obtain a corresponding probability density via the Boltzmann distribution

\begin{equation}
    p(\Xt) = \frac{1}{Z} \exp\!\left( -\frac{E(\Xt|\Xclean)}{k_\mathrm{B} T_{\mathrm{phys}}} \right) = \frac{1}{Z} \exp\!\left( -\frac{ \|\Xt - \Xclean\|_{F}^2}{k_\mathrm{B} T_{\mathrm{phys}}} \right)\,,
    \label{eq:boltzmann}
\end{equation}

where $k_\mathrm{B}$ is the Boltzmann constant and $Z$ is the normalizing partition function.
It assigns higher density to low-energy configurations, with the strength of this preference controlled by the physical temperature $T_{\mathrm{phys}}$.
The Boltzmann distribution of our GPFF is identical to the perturbation kernel in variance exploding diffusion (Eq.~\ref{eq:ve_kernel_one_step}) if we set the temperature depending on the noise level $T_{\mathrm{phys}}=2\sigma_t^2 \big/k_\mathrm{B}$.
In fact, we can directly reconstruct the score of the perturbation kernel (Eq.~\ref{eq:score_ve}) with the forces predicted by a GPFF just as with the noise direction predicted by a diffusion model:

\begin{equation}
    \spred(\Xt, t) 
    = -\frac{\epspred(\Xt, t)}{\sigma_{t}} 
    = \frac{\Xpred-\Xt}{\sigma_{t}^2} 
    = \frac{\Fpred(\Xt)}{2\sigma_{t}^2} \, .
    \label{eq:score_noise}
\end{equation}

Accordingly, GPFFs can be utilized with any existing sampling strategy for variance exploding diffusion models. However, there exist important differences between GPFFs and diffusion models in the training target and time-step handling.
While diffusion models are usually trained on quantities with unit variance such as the normalized noise direction $\eps_t$, the forces learned by GPFFs have a magnitude that implicitly encodes the noise level $\sigma_t$:

\begin{equation}
    \epspred(\Xt, t) \approx \frac{\Xt - \Xclean}{\sigma_t} = \eps_t
    \quad \text{and} \quad 
    \Fpred(\Xt) \approx -2(\Xt-\Xclean) = -2\sigma_t\eps_t \, .
    \label{eq:connection_training_objective}
\end{equation}

Furthermore, diffusion models are conditioned on the time step $t$ (or the noise level $\sigma_t$).
This means that they are limited to reverse processes with a fixed schedule where the time step and the corresponding noise level $\sigma_t$ are always known.
In contrast, GPFFs are trained without conditioning on $t$, they only take noisy structures $\Xt$ as input and implicitly estimate the noise level when predicting forces, which was shown to be feasible with Gaussian noise in a high dimensional space \cite{kahouli2024molecular}.
We confirm this empirically in Appendix~\ref{app:ablation_t}, where adding explicit time-step conditioning to GPFF brings no measurable improvement; the converse experiment, removing time conditioning from the DM, is reported in Appendix~\ref{app:dm_time_ablation} and shows a large drop in validity.
Therefore, they can be interpreted as time-step-agnostic diffusion models that enable denoising of arbitrary input structures and facilitate adaptive sampling strategies.

\subsection{Sampling strategies} \label{subsec:sampling_strategies}
We develop and evaluate several sampling algorithms that exploit the flexibility of GPFF as the second main contribution of this work.
Because GPFF is time-step-agnostic, denoising steps do not require a fixed noise schedule, allowing a wide range of sampling strategies and priors.
In the following, we propose direct denoising, a family of sampling approaches that iteratively refine a direct prediction of $\Xclean$ similar to fixed-point iteration schemes.
Furthermore, we show how GPFFs can be employed with traditional, diffusion-based samplers and then propose adaptive variants that specifically leverage the time-step-agnostic design of GPFFs.
Pseudocode for all algorithms is provided in Appendix~\ref{app:sampling_algorithms}.
Since some strategies take less denoising steps than the number of steps $T$ employed during training, we adopt a separate notation for the denoising steps: $i \in \{0, 1, \ldots, \Nsteps\}$ denotes the step index, where $\Xrev{0}$ corresponds to a sample from the prior and $\Nsteps$ is the maximum number of steps.

\paragraph{Direct denoising}
Our base strategy, direct denoising (DD), is inspired by structure relaxation with MLFFs and directly uses the predicted pseudo-forces $\Fpred$ to drive noisy geometries towards local minima in the pseudo-PES.
Because the pseudo-PES is locally quadratic, pseudo-forces admit a closed-form update for equilibrium coordinates from any point $\Xrev{i}$ close to a minimum:
\begin{align}
    \Xpred(\Xrev{i}) = \Xrev{i} + \frac{1}{2} \Fpred(\Xrev{i}) \label{eq:x0_prediction_gpff}
\end{align}
Structures with larger amounts of noise might have come from different reference equilibria and therefore the learned forces will rather point towards a region of less noisy configurations than directly to a single minimum.
Therefore, we design DD as an iterative sampler, where each step further refines the structure until it converges according to a force threshold $\|\Fpred\|_{\max} \leq f_{\max}$:
\begin{align}
    \Xrev{i+1} = \Xpred(\Xrev{i}) \quad .
    \label{eq:x0_model}
\end{align}
Similar to MLFF-based relaxation, sampling can start from a flexible choice of prior and intermediate structures can be modified intuitively.
We define three additional variants of DD.
First, a stochastic variant (SDD) where intermediate coordinates are perturbed with isotropic Gaussian noise, scaled by a linearly decaying schedule, allowing to explore more local minima:
\begin{equation}
    \Xrev{i+1} = \Xpred(\Xrev{i}) + \sigi{i} \cdot \epsi{i} \quad \text{with} \quad \epsi{i} \sim \gaussian \quad \text{and} \quad \sigi{i} = 1 - \frac{i}{\Nsteps} \quad .
\end{equation}
Second, we propose DD+shape for guiding the sampler towards a target shape parameterized by a $3\times3$ covariance matrix $\boldsymbol{\Sigma}_{\mathrm{target}}$, whose diagonal in the principal frame gives the target principal variances $\boldsymbol{\lambda}_{\mathrm{target}}=(\lambda_x, \lambda_y, \lambda_z)$.
Before each denoising step, the current positions are aligned to their respective principal frames and the current variances $\boldsymbol{\tilde{\lambda}}$ are rescaled by soft element-wise multiplication to match the target variances.
To avoid over-constraining the geometry near equilibrium, we interpolate between the corrected and uncorrected positions using $\alpha = (i/\Nsteps)^p$, where $p$ governs the strictness:
\begin{equation}
\Xrev{i}_{\mathrm{shaped}} = (1-\alpha) \cdot \sqrt{\frac{\boldsymbol{\lambda}_{\mathrm{target}}}{\boldsymbol{\tilde{\lambda}}}} \cdot \Xrev{i} + \alpha \cdot \Xrev{i} \quad .
\end{equation}
Finally, we propose DD+scaffold that enforces a valid molecular sub-structure during sampling.
\mbox{DD+scaffold} starts from a mixed prior $\Xrev{0}$ where some positions are defined by the target sub-structure and remaining atoms are sampled from Gaussian noise.
At every denoising step, the pseudo-forces on scaffold atoms are set to zero, while non-scaffold atoms are free to move.
All three variants can be combined with each other and with either DD or SDD.
However, SDD and DD+shape introduce $i$-dependent schedules for noise injection and shape enforcement respectively, which prevent the use of the force-based stopping criterion $f_{\max}$; all configurations involving these variants therefore use a fixed number of steps $\Nsteps$.

\paragraph{Diffusion-based adaptive sampling (AS)}
With the score-force conversion of Equation~\eqref{eq:score_noise}, GPFF predictions can be used directly inside standard diffusion samplers.
Unlike the DM, GPFF ignores the time-step input and relies solely on the geometry, but the fixed noise schedule still governs the step sizes and noise injection of the sampler.
In this work, we follow the EDM noise schedule by Karras~et~al. \cite{diff_edm}
\begin{equation}
    \sigi{i+1}(\sigi{i}, \Delta s) = \left( (\sigi{i})^{\frac{1}{\rho}} - \Delta s \cdot (\sigma_{\mathrm{max}}^{\frac{1}{\rho}} - \sigma_{\mathrm{min}}^{\frac{1}{\rho}})\right)^{\rho} \quad ,
    \label{eq:edm_schedule}
\end{equation}
where $\sigi{0}=\sigma_{\mathrm{max}}$, $\rho$ defines the steepness of the noise decay and the step size $\Delta s = \frac{1}{N-1}$ is determined by the number of steps and constant in $\rho$-space.

As the model prediction is approximate and the sampler discretizes a continuous process into finite steps, both the model and the sampler introduce errors during denoising.
These errors can cause the observed noise level $\sighat{i}$ to diverge from the schedule $\sigi{i}$.
Stochastic samplers partially compensate through noise injection, which limits error accumulation at the cost of additional NFE \cite{diff_edm, kahouli2025}.
However, because the pseudo-forces also encode the magnitude of the perturbation $\sigi{i} \epsi{i}$, Equation~\eqref{eq:pseudo_forces} naturally provides a cost-free estimate of the true noise level
\begin{equation}
    \sighat{i}=\mathrm{std}(\Fpred(\Xrev{i})/2) \quad .
    \label{noise_estimate}
\end{equation}
Comparing $\sighat{i}$ to the estimate at the previous step $\sighat{i-1}$ additionally yields an estimate of the true step size taken by the previous denoising step
\begin{equation}
    \Delta \hat{s}^{(i-1)} = \frac{{(\sighat{i-1}})^{\frac{1}{\rho}} - ({\sighat{i}})^{\frac{1}{\rho}}}{\sigma_{\mathrm{max}}^{\frac{1}{\rho}} - \sigma_{\mathrm{min}}^{\frac{1}{\rho}}} \quad .
\end{equation}
We leverage these estimates to adjust the schedule as follows.
At every step, AS replaces the scheduled noise level with the estimate $\sigi{i} = \sighat{i}$.
The sampler is initialized with an ambitious target step size $\Delta s^{(0)} = \frac{1}{\Ntarget-1}$, aiming for $\Ntarget\ll N$ steps.
From $i \geq 1$, the step size is additionally adjusted by adopting the observed value $\Delta s^{(i)} = \Delta \hat{s}^{(i-1)}$, and the next noise level is computed as $\sigma_{\mathrm{AS}}^{(i+1)} = \sigma^{(i+1)}(\sighat{i},\, \Delta \hat{s}^{(i-1)})$.
If the model denoises faster than $\Ntarget$ prescribes, the step size grows and sampling terminates early; if slower, the adjusted step size yields a more conservative schedule for subsequent steps.
To prevent stagnation, we define a non-adaptive upper bound $\sigma_{\mathrm{upper}}^{(i+1)}$ based on $\Nsteps$ steps, ensuring termination by enforcing $\sigma_{\mathrm{AS}}^{(i+1)} \leq \sigma_{\mathrm{upper}}^{(i+1)}$.
The mechanism applies identically to ancestral, Heun, and stochastic Heun sampling.
For the stochastic Heun variant, part of the injected noise variance is subtracted from $\sighat{i}$ to ensure the adaptation reflects genuine denoising progress (details in Appendix~\ref{app:as_trajectories}).
As with DD, the dynamic schedule of AS also allows using arbitrary priors where the initial noise level is unknown.
We show example adaptive sampling trajectories and their target and maximum noise schedules in Appendix Figure~\ref{sfig:figure_5}.

\section{Results}
\label{sec:results}

We train both GPFF and a DM baseline on the valid 123{,}569 QM9 molecules \cite{qm9}.
In all experiments we generated 10{,}000 molecules, initialized from an isotropic Gaussian prior with $\sigma_{\max} = 30.0$~\AA{}, except for DD+shape which uses a shaped Gaussian prior from the shape predictor.
Model and sampling parameters as well as noise level sampling according to Karras~et~al. \cite{diff_edm}, can be found in Appendix \ref{app:hyperparameters}, \ref{app:sampler_parameters}, and \ref{app:noise_levels}, respectively.

We assess the quality of the generated structures with a threefold test:
\begin{itemize}
    \item \textit{Validity} requires a parseable, connected, radical-free molecular graph as per RDKit's \texttt{DetermineBonds}\cite{rdkit}.
    \item \textit{Distributional agreement} between the valid molecules and QM9 is measured by Jensen--Shannon (JS) divergence (100-bin histograms) of maximum pairwise atomic distance (MPD), HOMO--LUMO gap, and atomization energy $U_0$ (latter two as per another PaiNN model, see Appendix~\ref{app:property_models}).
    \item \textit{Computational cost} is reported as the mean NFEs per molecule. Note that Heun-type samplers require $2N-1$ NFE for $N$ schedule steps, while ancestral sampling and GPFF direct denoising use one NFE per step. 
\end{itemize}

\subsection{Direct denoising leads to a spherical bias}\label{subsec:direct_denoising}

\begin{figure}
\centering
\includegraphics[width=1\linewidth]{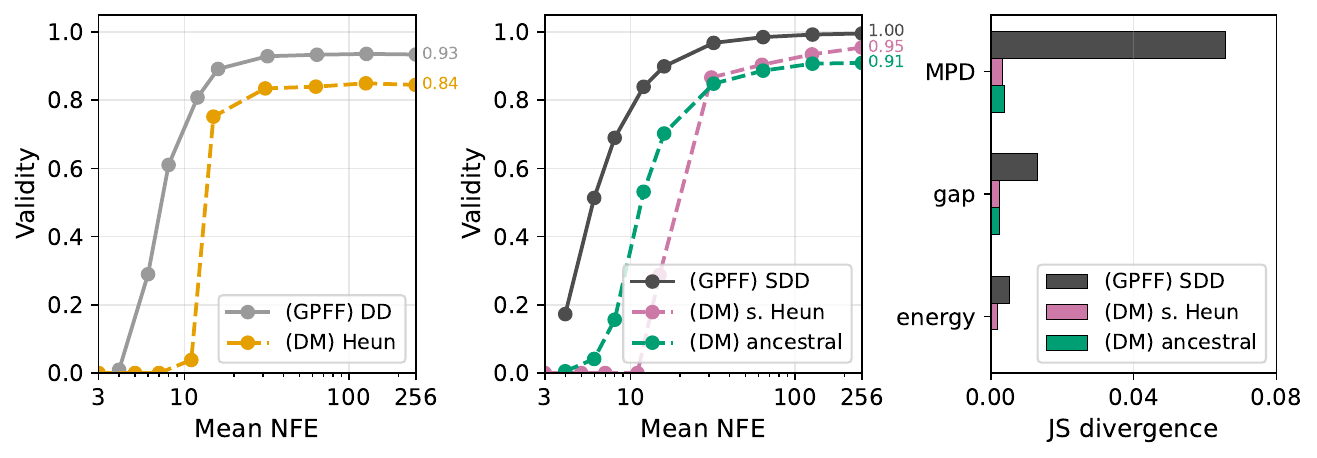}
\caption{\label{fig:figure_1} Comparison of GPFF (ours, solid) and DM baselines (dashed). 
Left: deterministic sampling, comparing GPFF with DD against the DM Heun sampler. 
Middle: stochastic sampling, comparing GPFF with DD against the DM stochastic Heun (s. Heun) and ancestral samplers.
Right: Jensen-Shannon (JS) divergence to QM9 at 250 NFE for the stochastic samplers, decomposed into MPD, HOMO--LUMO gap, and atomization energy contributions.}
\end{figure}

We evaluate GPFF with DD and SDD against the DM baseline using ancestral, Heun, and stochastic Heun sampling (see Figure~\ref{fig:figure_1}).
DD stops when reaching the $f_{\max}$ threshold values while
SDD and the diffusion-based samplers run for a fixed number of steps.
Our method GPFF consistently outperforms the DM baselines in validity:
in the deterministic regime GPFF reaches 93\% validity versus 84\% for Heun (left panel), and in the stochastic regime it achieves 100\% versus 95\% for stochastic Heun and 91\% for ancestral sampling.
Notably, GPFF already achieves approximately 50\% validity after only 6 NFE, while the DM baselines remain near 0\% at similar inference calls.

However, the JS divergence analysis evaluated at 256 NFEs (255 for Heun) reveals that GPFF's higher validity comes with a distributional bias (right panel).
Diffusion-based sampling achieves a low JS divergence in all three properties, while
GPFF exhibits higher JS divergence especially in MPD.
This effect stems from GPFF initially predicting a dataset-mean-like structure, which tends to be spherical.
In the next section, we introduce a two-fold strategy to alleviate this issue.

\subsection{Mitigating shape biases}\label{subsec:results_shape_bias}
To address the spherical bias from the previous section we introduce two complementary mitigation strategies:
First, we train a lightweight shape predictor (see Appendix~\ref{app:cov_predictor}) that, conditioned on the number of atoms, samples a target covariance $\boldsymbol{\Sigma}_{\mathrm{target}}$ used to draw the Gaussian prior; its principal variances $\boldsymbol{\lambda}_{\mathrm{target}}$ then enforce the target shape during SDD+shape sampling.
Second, instead of (S)DD, we employ diffusion-based samplers (see Section \ref{subsec:sampling_strategies}), where the model takes smaller, schedule-constrained steps that naturally prevent the trajectory from collapsing to compact geometries.

\begin{figure}
\centering
\includegraphics[width=1\linewidth]{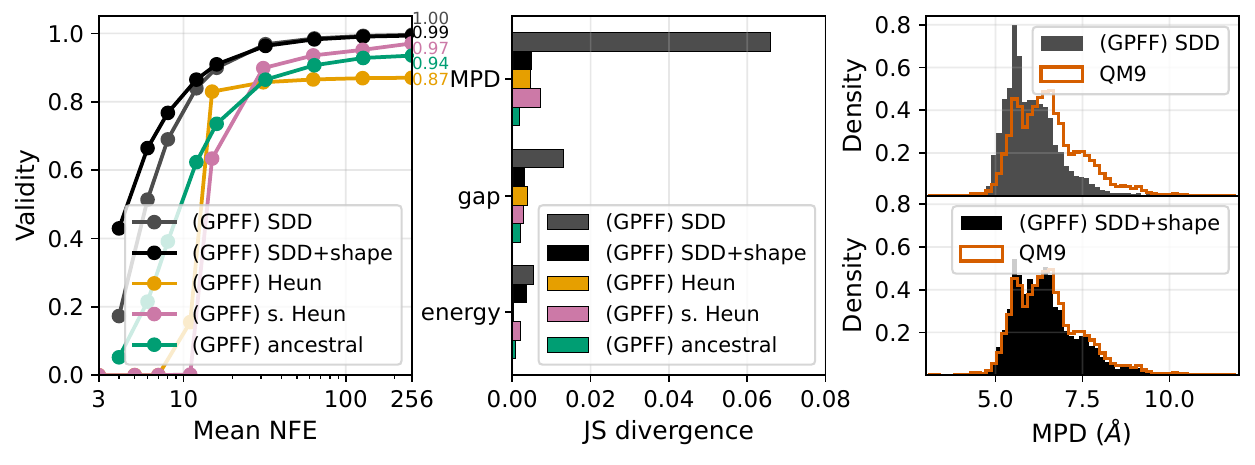}
\caption{\label{fig:figure_2}Alternative GPFF samplers and their similarity to QM9. Left: validity as a function of NFE for SDD, SDD+shape, Heun, stochastic Heun (s. Heun), and ancestral sampling. Middle: JS divergence to QM9 at 256 NFE for the same GPFF samplers, decomposed into MPD, gap, and energy contributions. Right: MPD-histograms of SDD, SDD+shape and QM9.}
\end{figure}

The results are shown in Figure~\ref{fig:figure_2}.
SDD+shape achieves the highest validity overall at 99\%, outperforming Heun (87\%), stochastic Heun (97\%), and ancestral sampling (94\%), with SDD from the previous section as comparison (100\%) (left panel).
The advantage is especially pronounced at low NFE: SDD+shape reaches approximately 40\% validity at just 4 NFE, surpassing even standard SDD at the same budget, because the variance-shaped prior already provides a starting point much closer to realistic molecular geometries.
All approaches successfully correct the spherical bias (middle panel): 
the JS divergence drops substantially compared to SDD, with ancestral sampling achieving marginally lower values, though the differences are small at these already low divergence levels.
The property histograms (right panel and Appendix~\ref{app:histograms}) further illustrate the effect.
SDD produces a pronounced peak at low MPD values compared to QM9, reflecting the spherical bias, while SDD+shape closely recovers the reference distribution.
This confirms that the bias is not a property of the GPFF model itself but of the (S)DD trajectory.
The combination of high validity, low NFE, and corrected distributional coverage makes DD+shape particularly suitable when both speed and distributional quality matter, and it is the basis of the targeted design experiments in Section~\ref{sec:targeted}.

\subsection{Exploiting GPFF's flexibility with adaptive sampling}\label{subsec:results_as}

\begin{figure}
\centering
\includegraphics[width=1\linewidth]{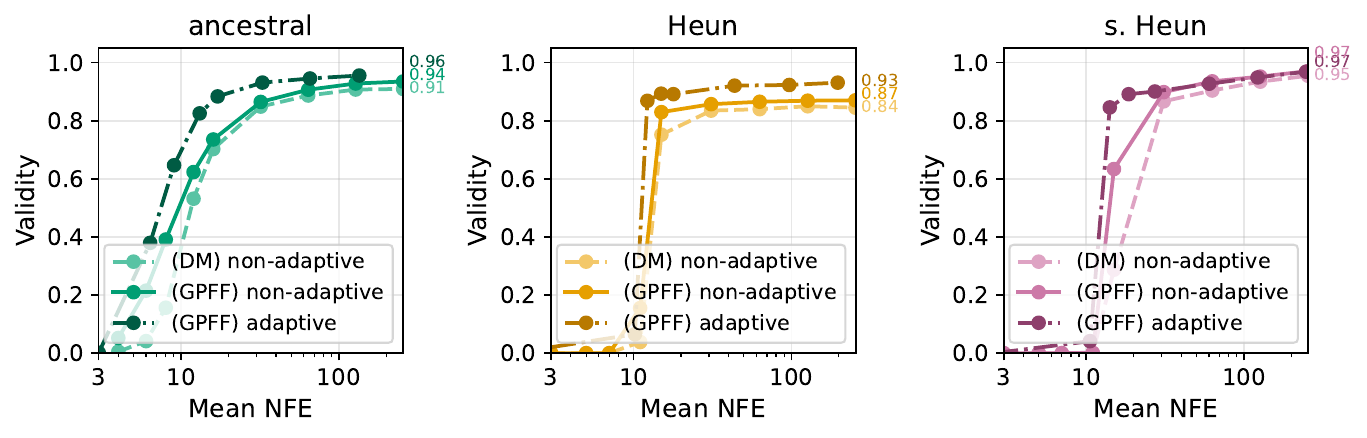}
\caption{\label{fig:figure_3}Adaptive versus non-adaptive sampling. Panels compare non-adaptive DM (dashed), non-adaptive GPFF (solid), and adaptive GPFF (dash-dotted) variants for ancestral sampling (left), Heun sampling (middle), and stochastic Heun sampling (right). 
Curves show validity as a function of NFE.}
\end{figure}

In the previous section, GPFF was used with fixed noise schedules for the diffusion-based samplers, leaving the step sizes unadjusted even when the model's own noise estimate suggested a different pace.
The AS mechanism described in Section~\ref{subsec:sampling_strategies} allows GPFF to dynamically adjust step sizes during the reverse trajectory.
To this end, we set an ambitious step target of $\Ntarget = N / 2$.

In Figure~\ref{fig:figure_3}, we show validity as a function of NFE for ancestral sampling (left panel), Heun sampling (middle panel) and stochastic Heun sampling (right panel), where AS outperforms its non-adaptive GPFF counterpart across all three sampler types, with the most improvement at low NFE.
At convergence, AS reaches 95\%, 92\% and 97\% validity for ancestral, Heun and stochastic Heun, compared to 94\%, 87\% and 97\% for its non-adaptive version.
Notably, even the non-adaptive GPFF samplers consistently outperform the DM baselines, achieving validities of 91\%, 84\% and 95\%, respectively.

This improved performance over DMs can be attributed to the model itself, because GPFF is inherently robust to deviations from the scheduled noise level: as a time-agnostic model, it does not receive the scheduled $\sigi{i}$ as input and instead responds to the actual geometry, making it implicitly adaptive by design.
The explicit step-size adjustment in AS amplifies this advantage, with the largest gains observed for the deterministic Heun sampler, where the corrective effect of noise injection is absent.

\subsection{Targeted molecular design}\label{sec:targeted}
Because GPFF is time-agnostic and does not require fixed diffusion schedules, it naturally supports sampling strategies that guide generation on a geometric level by modifying intermediate structures, analogous to what is possible with MLFFs.
In Section~\ref{subsec:results_shape_bias}, we showed that enforcing sampled target shapes during generation corrects the bias towards spherical molecules.
Here, we leverage this capability for targeted design: 
first generating molecules with prescribed geometric shapes using SDD+shape, and then adding scaffold constraints to fix substructures during generation.
Importantly, no additional GPFF model is trained for these experiments; we only add constraints to the SDD sampler.

We extend the shape predictor introduced in Section~\ref{subsec:results_shape_bias} with conditioning on the number of atoms and the target relative variances, enabling it to generate appropriately shaped anisotropic Gaussian priors.
These serve both as the initial prior and as the constraint enforced at each step of the DD+shape sampler (256 steps).
To demonstrate how we can freely steer the geometry towards a desired shape, we generate 10{,}000 for each rod-like, sphere-like, and disc-like molecules and plot them by their mass-free principal moments of inertia~\cite{sauer2003molecular} in 
Figure~\ref{fig:covariance_generation} (details in Appendix~\ref{app:pmi}).
In the left panel, the distribution of QM9 is shown, where the isolated cluster near the disc-rod edge corresponds to molecules whose heavy atoms are coplanar, with only hydrogen atoms contributing variance out of the plane.
The generated molecules (right panel) match the prescribed target shapes for all three geometries, with validity exceeding 98\% in each case.

\begin{figure}
\centering
\includegraphics[width=1\linewidth]{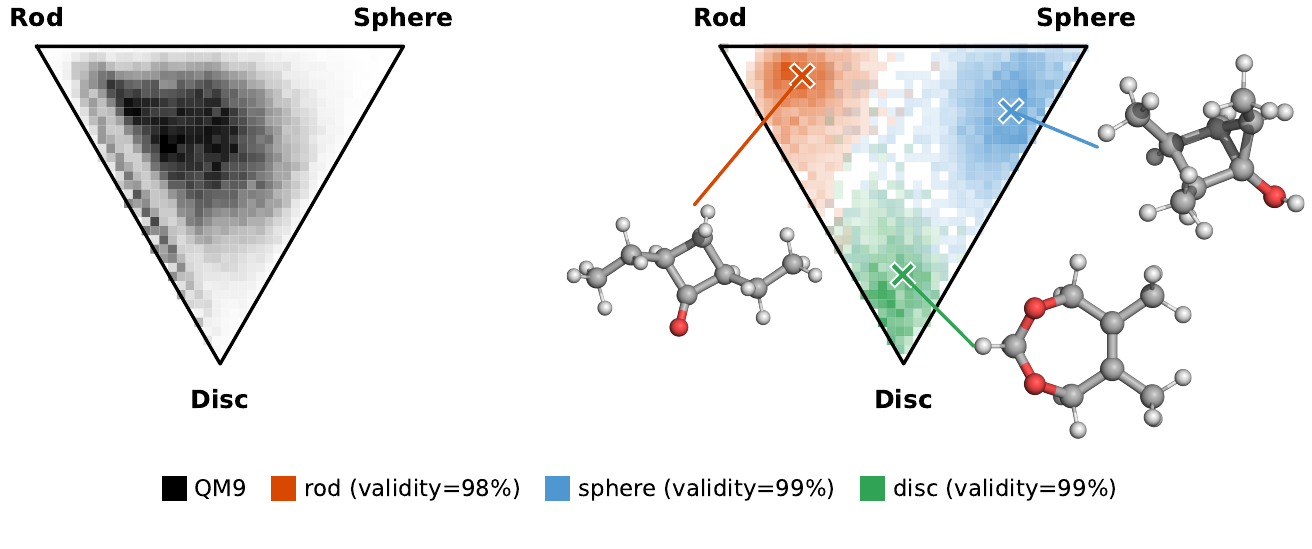}
\caption{\label{fig:covariance_generation}
Each point represents a molecule mapped to the shape space spanned by rod, sphere, and disc geometries according to its mass-free principal moments of inertia.
The enumerative QM9 data set shows a distribution across all three corners of the diagram (left). 
This flat distribution can be steered with SDD+shape as is shown in the right panel, where a sample of 10{,}000 structures was generated for each shape, demonstrating the flexibility of our approach.
Crosses mark the representative example molecules.
The generated clusters match their respective targets, with validity exceeding 98\% in all cases.}
\end{figure}

\begin{figure}
\centering
\includegraphics[width=0.95\linewidth]{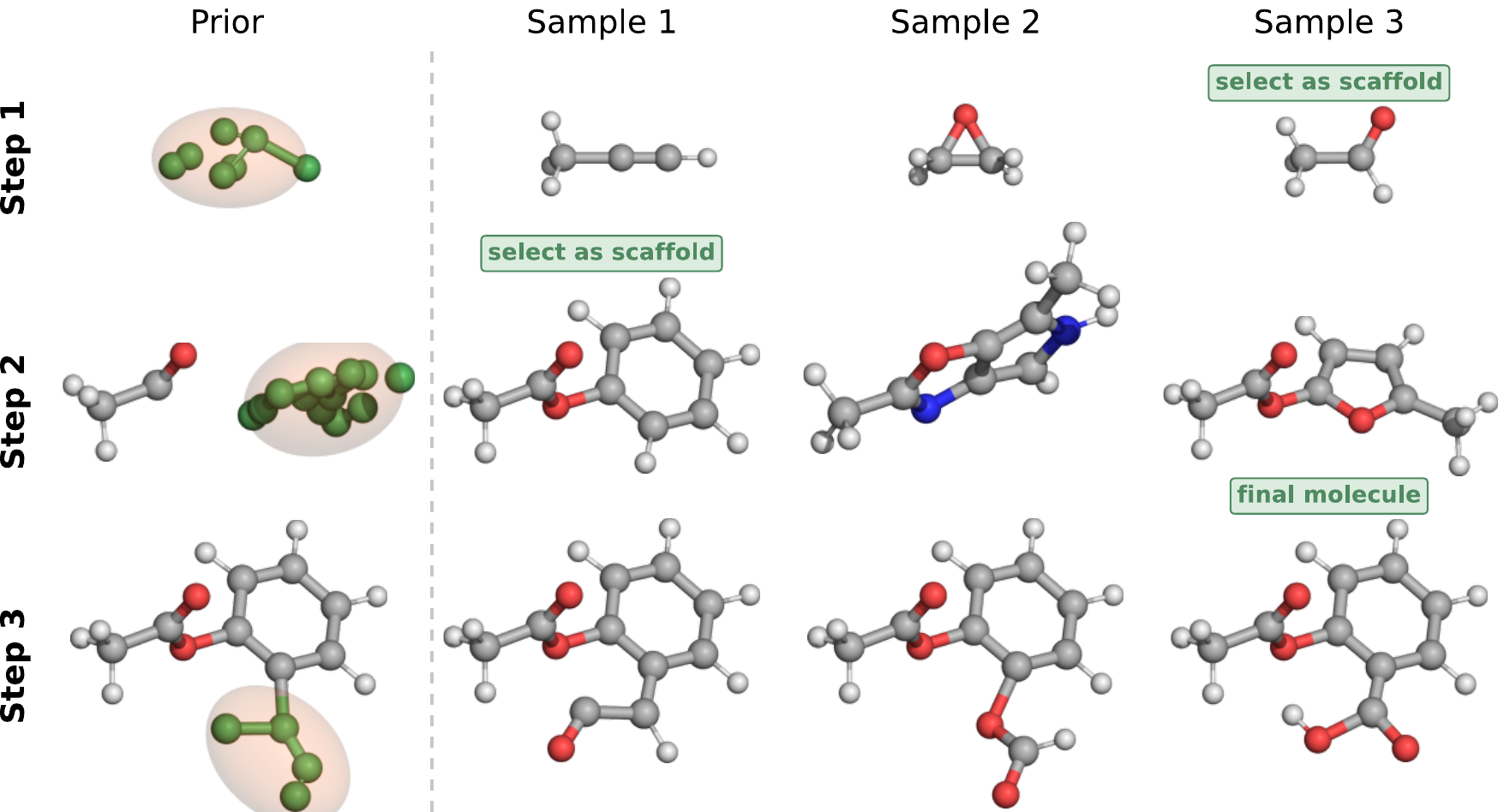}
\caption{\label{fig:molecule_paint}Iterative scaffold-guided design with shape-constrained denoising. Each row shows one stage of a three-step design process which recovers aspirin. 
In the left column, a prior (orange) is sampled around the current scaffold while keeping scaffold atoms fixed and the randomized atom types in green. The number of atoms and shape of each prior are chosen to steer the design toward aspirin. The remaining columns show representative denoised samples obtained from this prior distribution. One sample per row is highlighted as the selected structure passed to the next design stage.}
\end{figure}

To further illustrate the possibilities for constrained generation with GPFF, we build a simple iterative design tool that combines DD+shape with SDD+scaffold (see Figure~\ref{fig:molecule_paint}).
While this example serves as an illustrative proof of concept, the approach could be extended to realistic scenarios such as drug discovery, where a desired functional group must be placed within the geometric constraints of a protein pocket\cite{plainer2023diffdockpocket,schneuing2024structure,igashov2024equivariant,huang2024dual}.

We demonstrate the tool by iteratively assembling an out-of-distribution molecule with more heavy atoms than any molecule in QM9.
At every stage, the chemist can create a prior (orange blob, left column of Figure~\ref{fig:molecule_paint}) that defines the target covariance $\boldsymbol{\Sigma}_{\mathrm{target}}$ of the sampling distribution, with randomized atom types (denoted in green).
From this prior, three (or more) samples are drawn, in this case, propyne, ethylene oxide, and acetaldehyde.
We select one candidate as the new scaffold (removing one hydrogen atom), here acetaldehyde from the \ce{C2H4O} sample, and add the next prior (step 2) in order to extend the molecule to the right. Here, sample 1 with the \ce{C8H8O2} (phenyl acetate) atoms was selected.
In a last step, a four-atom prior is placed at the ortho position of the benzene. The selected sample contains \ce{CHO2} atoms, forming a carboxylic acid and recovering aspirin.
This demonstrates that the MLFF-like behavior of GPFF enables intuitive, iterative molecular design through geometric and structural constraints alone.

\section{Discussion and conclusion}
\label{sec:discussion_and_conclusion}
In this work, we introduced GPFF, a generative model that bridges machine learning force fields (MLFFs) and diffusion models (DMs) by reinterpreting the denoising process as relaxation on a pseudo potential energy surface.
GPFF adopts the data augmentation strategy of DMs but trains on pseudo-forces using a classical MLFF regression loss, bringing a new intuition to generative modeling that is more accessible to force-field-based reasoning.

A key novelty is that GPFF is time-step-agnostic, predicting pseudo-forces whose magnitude naturally scales with the perturbation level.
This dual nature allows GPFF to be used from two complementary perspectives.
From the MLFF perspective, pseudo-forces drive iterative structure relaxation towards equilibrium, enabling direct denoising (DD) and its variants: SDD (stochastic noise injection), DD+shape (geometric shape enforcement), and DD+scaffold (fixed sub-structures), which can be freely combined and support flexible priors beyond the standard isotropic Gaussian.
Following pseudo-forces serves as an attempt to recover the equilibrium geometry in a single step, taking the largest possible denoising update; being time-step-agnostic, GPFF can iteratively refine through repeated pseudo-force updates until convergence.
From the diffusion perspective, the score function can be derived directly from the pseudo-forces, allowing GPFF to serve as a drop-in replacement in standard diffusion samplers such as ancestral, Heun, or stochastic Heun sampling.
The magnitude-carrying prediction further enables cost-free noise-level estimation, which we exploit for adaptive sampling (AS) to dynamically correct for discretization and prediction errors.

Evaluated on QM9, SDD+shape achieves 99\% validity at 256 NFE and 70\% at 6 NFE, outperforming all diffusion-based alternatives, including stochastic Heun at 95\% and ancestral sampling below 10\% at 6 NFE.
GPFF also improves validity over the diffusion baseline across all diffusion-based samplers, which we attribute to its robustness against deviations from the noise schedule.
Adaptive sampling further improves performance, especially at low NFE budgets.
We also demonstrated targeted design capabilities by generating molecules with prescribed geometric shapes and iteratively assembling out-of-distribution structures with more heavy atoms than in training molecules through scaffold-guided denoising.
Overall, GPFF's time-step-agnostic design enables two complementary capabilities, both unavailable to standard diffusion models and outperforming them in validity: AS on top of any standard diffusion sampler, and DD with intuitive geometric and structural constraints during generation.
By framing generative modeling in the language of forces and structure relaxation, GPFF makes the field more accessible to the molecular simulation community.
This study evaluated GPFF exclusively on QM9, which contains small organic molecules with at most nine heavy atoms and its performance on larger systems remains to be demonstrated.
Furthermore, GPFF generates only atomic positions with a given composition and does not jointly generate atom types, limiting its applicability to settings where the molecular formula is unknown.

Several directions for future work emerge.
The pseudo-PES could be extended beyond the simple quadratic form, for instance by incorporating physical energy labels to bias the landscape towards chemically relevant minima.
The framework could be broadened to periodic systems, larger molecules, and diverse atom types.
The DD framework invites additional geometric and chemical constraints for application-specific design.
Finally, GPFF could serve as a pre-training stage for MLFFs, stabilizing them in high-energy regions where training data is typically scarce.

\section*{Data availability}
Code and data will be made available upon publication.

\section*{Acknowledgments}
SG was supported by the Postdoc.Mobility fellowship by the Swiss National Science Foundation (project no. 225476).
MP is supported by the Konrad Zuse School of Excellence in Learning and Intelligent Systems (\href{https://eliza.school/}{ELIZA}) through the DAAD programme Konrad Zuse Schools of Excellence in Artificial Intelligence, sponsored by the Federal Ministry of Education and Research.
K.R.M. was supported in part by the German Federal Ministry for Research, Technology and Space (BMFTR) under grants 01IS18025A, 031L0207D, 01IS18037A, 16IS24087C. K.R.M. was also supported by the Institute of Information \& communications Technology Planning \& Evaluation (IITP) grants funded by the Korea government (MSIT) (No. RS-2019-II190079, Artificial Intelligence Graduate School Program of Korea University and No. RS-2024-00457882, AI Research Hub Project).
We gratefully acknowledge support by the Deutsche Forschungsgemeinschaft (SFB1114, Projects No. A04 and No. B08) and the Berlin Mathematics center MATH+ (AA1-10, AA2-20, AA-Health-2).
During the preparation of this manuscript and during coding we used Claude Opus 4.6 and 4.7 for drafting and polishing purposes.
All outputs were reviewed and edited by the authors, who take full responsibility for the content.

\newpage
\bibliography{sample}

@article{chen2023importance,
  title={On the importance of noise scheduling for diffusion models},
  author={Chen, Ting},
  journal={arXiv preprint arXiv:2301.10972},
  year={2023}
}

@inproceedings{lin2024common,
  title={Common diffusion noise schedules and sample steps are flawed},
  author={Lin, Shanchuan and Liu, Bingchen and Li, Jiashi and Yang, Xiao},
  booktitle={Proceedings of the IEEE/CVF winter conference on applications of computer vision},
  pages={5404--5411},
  year={2024}
}

@inproceedings{
kim2024consistency,
title={Consistency Trajectory Models: Learning Probability Flow {ODE} Trajectory of Diffusion},
author={Kim, Dongjun and Lai, Chieh-Hsin and Liao, Wei-Hsiang and Murata, Naoki and Takida, Yuhta and Uesaka, Toshimitsu and He, Yutong and Mitsufuji, Yuki and Ermon, Stefano},
booktitle={The Twelfth International Conference on Learning Representations},
year={2024},
url={https://openreview.net/forum?id=ymjI8feDTD}
}

@misc{boffi2025buildconsistencymodellearning,
      title={How to build a consistency model: Learning flow maps via self-distillation}, 
      author={Boffi, Nicholas M. and Albergo, Michael S. and Vanden-Eijnden, Eric},
      year={2025},
      eprint={2505.18825},
      archivePrefix={arXiv},
      primaryClass={cs.LG},
      url={https://arxiv.org/abs/2505.18825}, 
}

@article{Arts2023,
	title        = {Two for One: Diffusion Models and Force Fields for Coarse-Grained Molecular Dynamics},
	author       = {Arts, Marloes and Garcia Satorras, Victor and Huang, Chin-Wei and Z{\"u}gner, Daniel and Federici, Marco and Clementi, Cecilia and No{\'e}, Frank and Pinsler, Robert and van den Berg, Rianne},
	year         = {2023},
	month        = {Sep},
	day          = {26},
	journal      = {Journal of Chemical Theory and Computation},
	publisher    = {American Chemical Society},
	volume       = {19},
	number       = {18},
	pages        = {6151--6159},
	doi          = {10.1021/acs.jctc.3c00702},
	issn         = {1549-9618},
	url          = {https://doi.org/10.1021/acs.jctc.3c00702}
}

@article{behler2007generalized,
	title        = {Generalized neural-network representation of high-dimensional potential-energy surfaces},
	author       = {Behler, J{\"o}rg and Parrinello, Michele},
	year         = {2007},
	journal      = {Physical Review Letters},
	publisher    = {APS},
	volume       = {98},
	number       = {14},
	pages        = {146401},
	doi          = {10.1103/physrevlett.98.146401}
}

@article{Bennett2023,
	title        = {Improving de novo protein binder design with deep learning},
	author       = {Bennett, Nathaniel R. and Coventry, Brian and Goreshnik, Inna and Huang, Buwei and Allen, Aza and Vafeados, Dionne and Peng, Ying Po and Dauparas, Justas and Baek, Minkyung and Stewart, Lance and DiMaio, Frank and De Munck, Steven and Savvides, Savvas N. and Baker, David},
	year         = {2023},
	month        = may,
	journal      = {Nature Communications},
	publisher    = {Springer Science and Business Media LLC},
	volume       = {14},
	number       = {1},
	doi          = {10.1038/s41467-023-38328-5},
	issn         = {2041-1723},
	url          = {http://dx.doi.org/10.1038/s41467-023-38328-5}
}

@article{Bhowmik2019inverse_batteries,
	title        = {A perspective on inverse design of battery interphases using multi-scale modelling, experiments and generative deep learning},
	author       = {Bhowmik, Arghya and Castelli, Ivano E. and Garcia-Lastra, Juan Maria and J{\o}rgensen, Peter Bj{\o}rn and Winther, Ole and Vegge, Tejs},
	year         = {2019},
	journal      = {Energy Storage Materials},
	volume       = {21},
	pages        = {446--456},
	doi          = {10.1016/j.ensm.2019.06.011},
	issn         = {2405-8297}
}

@article{bioemu,
	title        = {Scalable emulation of protein equilibrium ensembles with generative deep learning},
	author       = {Lewis, Sarah and Hempel, Tim and Jim{\'e}nez-Luna, Jos{\'e} and Gastegger, Michael and Xie, Yu and Foong, Andrew YK and Satorras, Victor Garc{\'\i}a and Abdin, Osama and Veeling, Bastiaan S. and Zaporozhets, Iryna and Chen, Yaoyi and Yang, Soojung and Foster, Adam E. and Schneuing, Arne and Nigam, Jigyasa and Barbero, Federico and Stimper, Vincent and Campbell, Andrew and Yim, Jason and Lienen, Marten and Shi, Yu and Zheng, Shuxin and Schulz, Hannes and Munir, Usman and Sordillo, Roberto and Tomioka, Ryota and Clementi, Cecilia and No{\'e}, Frank},
	year         = {2025},
	journal      = {Science},
	publisher    = {American Association for the Advancement of Science},
	pages        = {eadv9817},
	doi          = {10.1126/science.adv9817}
}

@book{bishopPRML,
	title        = {Pattern recognition and machine learning},
	author       = {Bishop, Christopher M.},
	year         = {2006},
	publisher    = {Springer New York},
	series       = {Information Science and Statistics},
	url          = {https://www.microsoft.com/en-us/research/publication/pattern-recognition-machine-learning/}
}

@article{cgschnet,
	title        = {Inverse design of 3d molecular structures with conditional generative neural networks},
	author       = {Gebauer, Niklas W. A. and Gastegger, Michael and Hessmann, Stefaan S. P. and M{\"u}ller, Klaus-Robert and Sch{\"u}tt, Kristof T.},
	year         = {2022},
	journal      = {Nature Communications},
	volume       = {13},
	number       = {1},
	pages        = {973},
	doi          = {10.1038/s41467-022-28526-y},
	issn         = {2041-1723}
}

@article{Chmiela2018Towards,
	title        = {Towards exact molecular dynamics simulations with machine-learned force fields},
	author       = {Chmiela, Stefan and Sauceda, Huziel E. and M{\"u}ller, Klaus-Robert and Tkatchenko, Alexandre},
	year         = {2018},
	journal      = {Nature Communications},
	publisher    = {Nature Publishing Group},
	volume       = {9},
	number       = {1},
	pages        = {3887},
	doi          = {10.1038/s41467-018-06169-2}
}

@book{cramer2004,
	title        = {Essentials of {{Computational Chemistry}}: {{Theories}} and {{Models}}},
	shorttitle   = {Essentials of {{Computational Chemistry}}},
	author       = {Cramer, Christopher J.},
	year         = {2004},
	location     = {{Chichester}},
	publisher    = {{Wiley}},
	isbn         = {978-0-470-09182-1},
	edition      = {2},
	langid       = {english},
	pagetotal    = {618}
}

@inproceedings{DDPM_Ho,
	title        = {Denoising Diffusion Probabilistic Models},
	author       = {Ho, Jonathan and Jain, Ajay and Abbeel, Pieter},
	year         = {2020},
	booktitle    = {Advances in Neural Information Processing Systems},
	publisher    = {Curran Associates, Inc.},
	volume       = {33},
	pages        = {6840--6851},
	url          = {https://proceedings.neurips.cc/paper_files/paper/2020/file/4c5bcfec8584af0d967f1ab10179ca4b-Paper.pdf},
	editor       = {Larochelle, H. and Ranzato, M. and Hadsell, R. and Balcan, M.F. and Lin, H.}
}

@article{DFT,
	title        = {Inhomogeneous electron gas},
	author       = {Hohenberg, Pierre and Kohn, Walter},
	year         = {1964},
	journal      = {Physical Review},
	publisher    = {APS},
	volume       = {136},
	number       = {3B},
	pages        = {B864},
	doi          = {10.1103/PhysRev.136.B864}
}

@inproceedings{diff_edm,
	title        = {Elucidating the Design Space of Diffusion-Based Generative Models},
	author       = {Karras, Tero and Aittala, Miika and Aila, Timo and Laine, Samuli},
	year         = {2022},
	booktitle    = {Advances in Neural Information Processing Systems},
	url          = {https://openreview.net/forum?id=k7FuTOWMOc7},
	editor       = {Oh, Alice H. and Agarwal, Alekh and Belgrave, Danielle and Cho, Kyunghyun}
}

@inproceedings{diffbridges,
	title        = {Diffusion-based Molecule Generation with Informative Prior Bridges},
	author       = {Wu, Lemeng and Gong, Chengyue and Liu, Xingchao and Ye, Mao and Liu, Qiang},
	year         = {2022},
	booktitle    = {Advances in Neural Information Processing Systems},
	publisher    = {Curran Associates, Inc.},
	volume       = {35},
	pages        = {36533--36545},
	url          = {https://proceedings.neurips.cc/paper_files/paper/2022/file/eccc6e11878857e87ec7dd109eaa9eeb-Paper-Conference.pdf},
	editor       = {Koyejo, S. and Mohamed, S. and Agarwal, A. and Belgrave, D. and Cho, K. and Oh, A.}
}

@inproceedings{edm,
	title        = {Equivariant Diffusion for Molecule Generation in 3{D}},
	author       = {Hoogeboom, Emiel and Satorras, V\'{\i}ctor Garcia and Vignac, Cl{\'e}ment and Welling, Max},
	year         = {2022},
	month        = {17--23 Jul},
	booktitle    = {Proceedings of the 39th International Conference on Machine Learning},
	publisher    = {PMLR},
	series       = {Proceedings of Machine Learning Research},
	volume       = {162},
	pages        = {8867--8887},
	url          = {https://proceedings.mlr.press/v162/hoogeboom22a.html},
	editor       = {Chaudhuri, Kamalika and Jegelka, Stefanie and Song, Le and Szepesvari, Csaba and Niu, Gang and Sabato, Sivan}
}

@article{efron2011tweedie,
	title        = {Tweedie’s formula and selection bias},
	author       = {Efron, Bradley},
	year         = {2011},
	journal      = {Journal of the American Statistical Association},
	publisher    = {Taylor \& Francis},
	volume       = {106},
	number       = {496},
	pages        = {1602--1614}
}

@article{eissler2026simple,
	title        = {How simple can you go? an off-the-shelf transformer approach to molecular dynamics},
	author       = {Eissler, Max and Korjakow, Tim and Ganscha, Stefan and Unke, Oliver T. and M{\"u}ller, Klaus-Robert and Gugler, Stefan},
	year         = {2026},
	journal      = {The Journal of Chemical Physics},
	publisher    = {AIP Publishing},
	volume       = {164},
	number       = {9}
}

@article{frank2024euclidean,
	title        = {A Euclidean transformer for fast and stable machine learned force fields},
	author       = {Frank, J. Thorben and Unke, Oliver T. and M{\"u}ller, Klaus-Robert and Chmiela, Stefan},
	year         = {2024},
	journal      = {Nature Communications},
	publisher    = {Nature Publishing Group UK London},
	volume       = {15},
	number       = {1},
	pages        = {6539}
}

@article{frank2026sampling,
  title={Sampling 3d molecular conformers with diffusion transformers},
  author={Frank, J. Thorben and Ripken, Winfried and Lied, Gregor and M{\"u}ller, Klaus-Robert and Unke, Oliver T. and Chmiela, Stefan},
  journal={Advances in Neural Information Processing Systems},
  volume={38},
  pages={168881--168931},
  year={2026}
}

@article{frank2026machine,
	title        = {Machine learning global atomic representations with euclidean fast attention},
	author       = {Frank, J. Thorben and Chmiela, Stefan and M{\"u}ller, Klaus-Robert and Unke, Oliver T.},
	year         = {2026},
	journal      = {Nature Machine Intelligence},
	publisher    = {Nature Publishing Group UK London},
	volume       = {8},
	number       = {3},
	pages        = {388--402}
}

@article{Freeze2019inverse_catalysts,
	title        = {Search for Catalysts by Inverse Design: Artificial Intelligence, Mountain Climbers, and Alchemists},
	author       = {Freeze, Jessica G. and Kelly, H. Ray and Batista, Victor S.},
	year         = {2019},
	journal      = {Chemical Reviews},
	volume       = {119},
	number       = {11},
	pages        = {6595--6612},
	doi          = {10.1021/acs.chemrev.8b00759},
	note         = {PMID: 31059236},
	eprint       = {https://doi.org/10.1021/acs.chemrev.8b00759}
}

@article{ganscha2025qcml,
  title={The QCML dataset, Quantum chemistry reference data from 33.5 M DFT and 14.7 B semi-empirical calculations},
  author={Ganscha, Stefan and Unke, Oliver T. and Ahlin, Daniel and Maennel, Hartmut and Kashubin, Sergii and M{\"u}ller, Klaus-Robert},
  journal={Scientific Data},
  volume={12},
  number={1},
  pages={406},
  year={2025},
  publisher={Nature Publishing Group UK London}
}

@article{Gantzer2020inverse_drugs,
	title        = {Inverse-QSPR for de novo Design: A Review},
	author       = {Gantzer, Philippe and Creton, Benoit and Nieto-Draghi, Carlos},
	year         = {2020},
	journal      = {Molecular Informatics},
	volume       = {39},
	number       = {4},
	pages        = {1900087},
	doi          = {10.1002/minf.201900087},
	eprint       = {https://onlinelibrary.wiley.com/doi/pdf/10.1002/minf.201900087}
}

@inproceedings{garcia2021en_flow,
	title        = {E(n) Equivariant Normalizing Flows},
	author       = {Garcia Satorras, Victor and Hoogeboom, Emiel and Fuchs, Fabian and Posner, Ingmar and Welling, Max},
	year         = {2021},
	booktitle    = {Advances in Neural Information Processing Systems},
	publisher    = {Curran Associates, Inc.},
	volume       = {34},
	pages        = {4181--4192},
	url          = {https://proceedings.neurips.cc/paper_files/paper/2021/file/21b5680d80f75a616096f2e791affac6-Paper.pdf},
	editor       = {Ranzato, M. and Beygelzimer, A. and Dauphin, Y. and Liang, P.S. and Vaughan, J. Wortman}
}

@article{gdml,
	title        = {Machine Learning of Accurate Energy-Conserving Molecular Force Fields},
	author       = {Chmiela, Stefan and Tkatchenko, Alexandre and Sauceda, Huziel E. and Poltavsky, Igor and Sch{\"u}tt, Kristof T. and M{\"u}ller, Klaus-Robert},
	year         = {2017},
	journal      = {Science Advances},
	volume       = {3},
	number       = {5},
	pages        = {e1603015},
	doi          = {10.1126/sciadv.1603015}
}

@inproceedings{geodiff,
	title        = {{GeoDiff}: A Geometric Diffusion Model for Molecular Conformation Generation},
	author       = {Xu, Minkai and Yu, Lantao and Song, Yang and Shi, Chence and Ermon, Stefano and Tang, Jian},
	year         = {2022},
	booktitle    = {International Conference on Learning Representations},
	url          = {https://openreview.net/forum?id=PzcvxEMzvQC}
}

@inproceedings{geoldm,
	title        = {Geometric Latent Diffusion Models for 3{D} Molecule Generation},
	author       = {Xu, Minkai and Powers, Alexander S. and Dror, Ron O. and Ermon, Stefano and Leskovec, Jure},
	year         = {2023},
	month        = {23--29 Jul},
	booktitle    = {Proceedings of the 40th International Conference on Machine Learning},
	publisher    = {PMLR},
	series       = {Proceedings of Machine Learning Research},
	volume       = {202},
	pages        = {38592--38610},
	url          = {https://proceedings.mlr.press/v202/xu23n.html},
	editor       = {Krause, Andreas and Brunskill, Emma and Cho, Kyunghyun and Engelhardt, Barbara and Sabato, Sivan and Scarlett, Jonathan}
}

@inproceedings{gschnet,
	title        = {Symmetry-adapted generation of 3d point sets for the targeted discovery of molecules},
	author       = {Gebauer, Niklas W. A. and Gastegger, Michael and Sch{\"u}tt, Kristof T.},
	year         = {2019},
	booktitle    = {Advances in Neural Information Processing Systems},
	publisher    = {Curran Associates, Inc.},
	volume       = {32},
	pages        = {7566--7578},
	url          = {https://proceedings.neurips.cc/paper_files/paper/2019/file/a4d8e2a7e0d0c102339f97716d2fdfb6-Paper.pdf},
	editor       = {Wallach, H. and Larochelle, H. and Beygelzimer, A. and d\textquotesingle Alch{\'e}-Buc, F. and Fox, E. and Garnett, R.}
}

@article{gugler2020,
	title        = {Enumeration of de novo inorganic complexes for chemical discovery and machine learning},
	author       = {Gugler, Stefan and Janet, Jon Paul and Kulik, Heather J.},
	year         = {2020},
	journal      = {Molecular Systems Design \& Engineering},
	publisher    = {Royal Society of Chemistry (RSC)},
	volume       = {5},
	number       = {1},
	pages        = {139–152},
	doi          = {10.1039/c9me00069k},
	issn         = {2058-9689},
	url          = {http://dx.doi.org/10.1039/C9ME00069K}
}

@misc{gugler2025,
	title        = {Molecular Similarity in Machine Learning of Energies in Chemical Reaction Networks},
	author       = {Gugler, Stefan and Reiher, Markus},
	year         = {2025},
	publisher    = {arXiv},
	doi          = {10.48550/ARXIV.2504.18742},
	url          = {https://arxiv.org/abs/2504.18742},
	copyright    = {arXiv.org perpetual,  non-exclusive license}
}

@article{hajduk2007decade_drug_fragments,
	title        = {A decade of fragment-based drug design: strategic advances and lessons learned},
	author       = {Hajduk, Philip J. and Greer, Jonathan},
	year         = {2007},
	journal      = {Nature Reviews Drug Discovery},
	publisher    = {Nature Publishing Group},
	volume       = {6},
	number       = {3},
	pages        = {211--219},
	doi          = {10.1038/nrd2220}
}

@article{hautier2011novel_batteries_throughput,
	title        = {Novel mixed polyanions lithium-ion battery cathode materials predicted by high-throughput ab initio computations},
	author       = {Hautier, Geoffroy and Jain, Anubhav and Chen, Hailong and Moore, Charles and Ong, Shyue Ping and Ceder, Gerbrand},
	year         = {2011},
	journal      = {Journal of Materials Chemistry},
	publisher    = {Royal Society of Chemistry},
	volume       = {21},
	number       = {43},
	pages        = {17147--17153},
	doi          = {10.1039/c1jm12216a}
}

@article{huang2024dual,
	title        = {A dual diffusion model enables 3D molecule generation and lead optimization based on target pockets},
	author       = {Huang, Lei and Xu, Tingyang and Yu, Yang and Zhao, Peilin and Chen, Xingjian and Han, Jing and Xie, Zhi and Li, Hailong and Zhong, Wenge and Wong, Ka-Chun and others},
	year         = {2024},
	journal      = {Nature Communications},
	publisher    = {Nature Publishing Group UK London},
	volume       = {15},
	number       = {1},
	pages        = {2657}
}

@article{igashov2024equivariant,
	title        = {Equivariant 3D-conditional diffusion model for molecular linker design},
	author       = {Igashov, Ilia and St{\"a}rk, Hannes and Vignac, Cl{\'e}ment and Schneuing, Arne and Satorras, Victor Garcia and Frossard, Pascal and Welling, Max and Bronstein, Michael and Correia, Bruno},
	year         = {2024},
	journal      = {Nature Machine Intelligence},
	publisher    = {Nature Publishing Group UK London},
	volume       = {6},
	number       = {4},
	pages        = {417--427}
}

@inproceedings{imp_ddpm,
	title        = {Improved Denoising Diffusion Probabilistic Models},
	author       = {Nichol, Alexander Quinn and Dhariwal, Prafulla},
	year         = {2021},
	month        = {18--24 Jul},
	booktitle    = {Proceedings of the 38th International Conference on Machine Learning},
	publisher    = {PMLR},
	series       = {Proceedings of Machine Learning Research},
	volume       = {139},
	pages        = {8162--8171},
	url          = {https://proceedings.mlr.press/v139/nichol21a.html},
	editor       = {Meila, Marina and Zhang, Tong}
}

@book{jensen2017b,
	title        = {Introduction to {{Computational Chemistry}}},
	author       = {Jensen, Frank},
	year         = {2017},
	location     = {{Chichester, UK ; Hoboken, NJ}},
	publisher    = {{Wiley}},
	isbn         = {978-1-118-82599-0},
	edition      = {3},
	langid       = {english},
	pagetotal    = {660}
}

@article{Joshi20213D,
	title        = {{3D-Scaffold}: A Deep Learning Framework to Generate 3D Coordinates of Drug-like Molecules with Desired Scaffolds},
	author       = {Joshi, Rajendra P. and Gebauer, Niklas W. A. and Bontha, Mridula and Khazaieli, Mercedeh and James, Rhema M. and Brown, James B. and Kumar, Neeraj},
	year         = {2021},
	journal      = {Journal of Physical Chemistry B},
	volume       = {125},
	number       = {44},
	pages        = {12166--12176},
	doi          = {10.1021/acs.jpcb.1c06437}
}

@article{kahouli2024molecular,
	title        = {Molecular relaxation by reverse diffusion with time step prediction},
	author       = {Kahouli, Khaled and Hessmann, Stefaan S. P. and M{\"u}ller, Klaus-Robert and Nakajima, Shinichi and Gugler, Stefan and Gebauer, Niklas W. A.},
	year         = {2024},
	journal      = {Machine Learning: Science and Technology},
	publisher    = {IOP Publishing},
	volume       = {5},
	number       = {3},
	pages        = {035038}
}

@misc{kahouli2025,
	title        = {Disentangling Total-Variance and Signal-to-Noise-Ratio Improves Diffusion Models},
	author       = {Kahouli, Khaled and Ripken, Winfried and Gugler, Stefan and Unke, Oliver T. and M{\"u}ller, Klaus-Robert and Nakajima, Shinichi},
	year         = {2025},
	publisher    = {arXiv},
	doi          = {10.48550/ARXIV.2502.08598},
	url          = {https://arxiv.org/abs/2502.08598},
	copyright    = {arXiv.org perpetual,  non-exclusive license}
}

@article{keith2021combining,
	title        = {Combining machine learning and computational chemistry for predictive insights into chemical systems},
	author       = {Keith, John A. and Vassilev-Galindo, Valentin and Cheng, Bingqing and Chmiela, Stefan and Gastegger, Michael and M{\"u}ller, Klaus-Robert and Tkatchenko, Alexandre},
	year         = {2021},
	journal      = {Chemical Reviews},
	publisher    = {ACS Publications},
	volume       = {121},
	number       = {16},
	pages        = {9816--9872}
}

@article{klein2023equivariant,
	title        = {Equivariant flow matching},
	author       = {Klein, Leon and Kr{\"a}mer, Andreas and No{\'e}, Frank},
	year         = {2023},
	journal      = {Advances in Neural Information Processing Systems},
	volume       = {36},
	pages        = {59886--59910}
}

@inproceedings{klein2023timewarp,
	title        = {Timewarp: Transferable Acceleration of Molecular Dynamics by Learning Time-Coarsened Dynamics},
	author       = {Klein, Leon and Foong, Andrew Y. K. and Fjelde, Tor Erlend and Mlodozeniec, Bruno Kacper and Brockschmidt, Marc and Nowozin, Sebastian and No{\'e}, Frank and Tomioka, Ryota},
	year         = {2023},
	booktitle    = {Thirty-seventh Conference on Neural Information Processing Systems},
	url          = {https://openreview.net/forum?id=EjMLpTgvKH}
}

@inproceedings{kohler2020equivariant,
	title        = {Equivariant Flows: Exact Likelihood Generative Learning for Symmetric Densities},
	author       = {K{\"o}hler, Jonas and Klein, Leon and No{\'e}, Frank},
	year         = {2020},
	month        = {13--18 Jul},
	booktitle    = {Proceedings of the 37th International Conference on Machine Learning},
	publisher    = {PMLR},
	series       = {Proceedings of Machine Learning Research},
	volume       = {119},
	pages        = {5361--5370},
	url          = {https://proceedings.mlr.press/v119/kohler20a.html},
	editor       = {III, Hal Daum{\'e} and Singh, Aarti}
}

@article{luhman2021knowledge,
	title        = {Knowledge distillation in iterative generative models for improved sampling speed},
	author       = {Luhman, Eric and Luhman, Troy},
	year         = {2021},
	journal      = {arXiv preprint arXiv:2101.02388}
}

@article{mdm,
	title        = {MDM: Molecular Diffusion Model for 3D Molecule Generation},
	author       = {Huang, Lei and Zhang, Hengtong and Xu, Tingyang and Wong, Ka-Chun},
	year         = {2023},
	month        = {Jun.},
	journal      = {Proceedings of the AAAI Conference on Artificial Intelligence},
	volume       = {37},
	number       = {4},
	pages        = {5105--5112},
	doi          = {10.1609/aaai.v37i4.25639},
	url          = {https://ojs.aaai.org/index.php/AAAI/article/view/25639}
}

@inproceedings{moldiff,
	title        = {{M}ol{D}iff: Addressing the Atom-Bond Inconsistency Problem in 3{D} Molecule Diffusion Generation},
	author       = {Peng, Xingang and Guan, Jiaqi and Liu, Qiang and Ma, Jianzhu},
	year         = {2023},
	month        = {23--29 Jul},
	booktitle    = {Proceedings of the 40th International Conference on Machine Learning},
	publisher    = {PMLR},
	series       = {Proceedings of Machine Learning Research},
	volume       = {202},
	pages        = {27611--27629},
	url          = {https://proceedings.mlr.press/v202/peng23b.html},
	editor       = {Krause, Andreas and Brunskill, Emma and Cho, Kyunghyun and Engelhardt, Barbara and Sabato, Sivan and Scarlett, Jonathan}
}

@article{montavon2013machine,
	title        = {Machine learning of molecular electronic properties in chemical compound space},
	author       = {Montavon, Gr{\'e}goire and Rupp, Matthias and Gobre, Vivekanand and Vazquez-Mayagoitia, Alvaro and Hansen, Katja and Tkatchenko, Alexandre and M{\"u}ller, Klaus-Robert and von Lilienfeld, O. Anatole},
	year         = {2013},
	journal      = {New Journal of Physics},
	publisher    = {IOP Publishing},
	volume       = {15},
	number       = {9},
	pages        = {095003}
}

@article{nequip,
	title        = {E(3)-equivariant graph neural networks for data-efficient and accurate interatomic potentials},
	author       = {Batzner, Simon and Musaelian, Albert and Sun, Lixin and Geiger, Mario and Mailoa, Jonathan P. and Kornbluth, Mordechai and Molinari, Nicola and Smidt, Tess E. and Kozinsky, Boris},
	year         = {2022},
	month        = {May},
	day          = {04},
	journal      = {Nature Communications},
	volume       = {13},
	number       = {1},
	pages        = {2453},
	doi          = {10.1038/s41467-022-29939-5},
	issn         = {2041-1723}
}

@article{noe2019boltzmann,
	title        = {Boltzmann generators: Sampling equilibrium states of many-body systems with deep learning},
	author       = {No{\'e}, Frank and Olsson, Simon and K{\"o}hler, Jonas and Wu, Hao},
	year         = {2019},
	journal      = {Science},
	publisher    = {American Association for the Advancement of Science},
	volume       = {365},
	number       = {6457},
	pages        = {eaaw1147},
	doi          = {10.1126/science.aaw1147}
}

@article{noe2020machine,
	title        = {Machine learning for molecular simulation},
	author       = {No{\'e}, Frank and Tkatchenko, Alexandre and M{\"u}ller, Klaus-Robert and Clementi, Cecilia},
	year         = {2020},
	journal      = {Annual Review of Physical Chemistry},
	publisher    = {Annual Reviews},
	volume       = {71},
	number       = {1},
	pages        = {361--390},
	doi          = {10.1146/annurev-physchem-042018-052331}
}

@inproceedings{painn,
	title        = {Equivariant message passing for the prediction of tensorial properties and molecular spectra},
	author       = {Sch{\"u}tt, Kristof T. and Unke, Oliver T. and Gastegger, Michael},
	year         = {2021},
	month        = {18--24 Jul},
	booktitle    = {Proceedings of the 38th International Conference on Machine Learning},
	publisher    = {PMLR},
	series       = {Proceedings of Machine Learning Research},
	volume       = {139},
	pages        = {9377--9388},
	url          = {https://proceedings.mlr.press/v139/schutt21a.html},
	editor       = {Meila, Marina and Zhang, Tong}
}

@article{physnet,
	title        = {{PhysNet}: a neural network for predicting energies, forces, dipole moments, and partial charges},
	author       = {Unke, Oliver T. and Meuwly, Markus},
	year         = {2019},
	journal      = {Journal of Chemical Theory and Computation},
	publisher    = {ACS Publications},
	volume       = {15},
	number       = {6},
	pages        = {3678--3693},
	doi          = {10.1021/acs.jctc.9b00181}
}

@inproceedings{Plainer2025,
  author = {Plainer, Michael and Wu, Hao and Klein, Leon and G{\"u}nnemann, Stephan and No{\'e}, Frank},
  title = {Consistent Sampling and Simulation: Molecular Dynamics with Energy-Based Diffusion Models},
  booktitle = {Advances in Neural Information Processing Systems},
  editor = {Belgrave, D. and Zhang, C. and Lin, H. and Pascanu, R. and Koniusz, P. and Ghassemi, M. and Chen, N.},
  pages = {24460--24505},
  publisher = {Curran Associates, Inc.},
  volume = {38},
  year = {2025},
}

@inproceedings{plainer2023diffdockpocket,
  author = {Plainer, Michael and Toth, Marcella and Dobers, Simon and St{\"a}rk, Hannes and Corso, Gabriele and Marquet, C{\'e}line and Barzilay, Regina},
  title = {{DiffDock-Pocket}: Diffusion for Pocket-Level Docking with Sidechain Flexibility},
  year = {2023},
  maintitle = {Advances in Neural Information Processing Systems},
  booktitle = {Machine Learning in Structural Biology},
}

@article{bonneau2026breaking,
  author = {Bonneau, Klara and Pasos-Trejo, Aldo S. and Plainer, Michael and Sagresti, Luca and Venturin, Jacopo and Zaporozhets, Iryna and Caruso, Alessandro and Rolando, Edoardo and Guljas, Andrea and Klein, Leon and Schebek, Maximilian and Albani, Filippo and L{\'o}pez-R{\'\i}os de Castro, Raquel and El Machachi, Zakariya and Giambagli, Lorenzo and Clementi, Cecilia},
  title = {Breaking the Barriers of Molecular Dynamics With Deep-Learning: Opportunities, Pitfalls, and How to Navigate Them},
  journal = {WIREs Computational Molecular Science},
  volume = {16},
  number = {1},
  pages = {e70064},
  year = {2026},
}

@article{Prot2010,
	title        = {Druggable pockets and binding site centric chemical space: a paradigm shift in drug discovery},
	author       = {P{\'e}rot, St{\'e}phanie and Sperandio, Olivier and Miteva, Maria A. and Camproux, Anne-Claude and Villoutreix, Bruno O.},
	year         = {2010},
	month        = aug,
	journal      = {Drug Discovery Today},
	publisher    = {Elsevier BV},
	volume       = {15},
	number       = {15–16},
	pages        = {656–667},
	doi          = {10.1016/j.drudis.2010.05.015},
	issn         = {1359-6446},
	url          = {http://dx.doi.org/10.1016/j.drudis.2010.05.015}
}

@article{qm7x,
	title        = {{QM7-X}, a comprehensive dataset of quantum-mechanical properties spanning the chemical space of small organic molecules},
	author       = {Hoja, Johannes and Medrano Sandonas, Leonardo and Ernst, Brian G. and Vazquez-Mayagoitia, Alvaro and DiStasio Jr., Robert A. and Tkatchenko, Alexandre},
	year         = {2021},
	month        = {Feb},
	day          = {02},
	journal      = {Scientific Data},
	volume       = {8},
	number       = {1},
	pages        = {43},
	doi          = {10.1038/s41597-021-00812-2},
	issn         = {2052-4463}
}

@article{qm9,
	title        = {Quantum chemistry structures and properties of 134 kilo molecules},
	author       = {Ramakrishnan, Raghunathan and Dral, Pavlo O. and Rupp, Matthias and von Lilienfeld, O. Anatole},
	year         = {2014},
	journal      = {Scientific Data},
	publisher    = {Nature Publishing Group},
	volume       = {1},
	number       = {1},
	pages        = {1--7}
}

@misc{rdkit,
	title        = {{RDKit}: Open-Source Cheminformatics},
	author       = {Landrum, Greg and Tosco, Paolo and Kelley, Brian and Rodriguez, Ricardo and Cosgrove, David and Vianello, Riccardo and sriniker and Gedeck, Peter and Jones, Gareth and Kawashima, Eisuke and Schneider, Nadine and Nealschneider, Dan and tadhurst-cdd and Dalke, Andrew and Swain, Matt and Cole, Brian and Turk, Samo and Savelev, Aleksandr and Maeder, Niels and Walker, Rachel and Vaucher, Alain and W{\'o}jcikowski, Maciej and Faara, Hussein and Take, Ichiru and Scalfani, Vincent F. and Probst, Daniel and Ujihara, Kazuya and Pechersky, Yakov and Monat, Jeremy and Lehtivarjo, Juuso},
	url          = {https://www.rdkit.org},
	note         = {DOI: 10.5281/zenodo.18797641}
}

@article{reymond2015,
	title        = {The Chemical Space Project},
	author       = {Reymond, Jean-Louis},
	year         = {2015},
	month        = feb,
	journal      = {Accounts of Chemical Research},
	publisher    = {American Chemical Society (ACS)},
	volume       = {48},
	number       = {3},
	pages        = {722–730},
	doi          = {10.1021/ar500432k},
	issn         = {1520-4898},
	url          = {http://dx.doi.org/10.1021/ar500432k}
}

@incollection{robbins1992empirical,
	title        = {An empirical Bayes approach to statistics},
	author       = {Robbins, Herbert E.},
	year         = {1992},
	booktitle    = {Breakthroughs in Statistics: Foundations and basic theory},
	publisher    = {Springer},
	pages        = {388--394}
}

@article{rupp2012fast,
	title        = {Fast and accurate modeling of molecular atomization energies with machine learning},
	author       = {Rupp, Matthias and Tkatchenko, Alexandre and M{\"u}ller, Klaus-Robert and von Lilienfeld, O. Anatole},
	year         = {2012},
	journal      = {Physical Review Letters},
	publisher    = {APS},
	volume       = {108},
	number       = {5},
	pages        = {058301}
}

@article{salimans2022progressive,
	title        = {Progressive distillation for fast sampling of diffusion models},
	author       = {Salimans, Tim and Ho, Jonathan},
	year         = {2022},
	journal      = {arXiv preprint arXiv:2202.00512}
}

@article{sauer2003molecular,
	title        = {Molecular shape diversity of combinatorial libraries: a prerequisite for broad bioactivity},
	author       = {Sauer, Wolfgang HB and Schwarz, Matthias K.},
	year         = {2003},
	journal      = {Journal of Chemical Information and Computer Sciences},
	publisher    = {ACS Publications},
	volume       = {43},
	number       = {3},
	pages        = {987--1003}
}

@article{schlegel2011,
	title        = {Geometry optimization},
	author       = {Schlegel, H. Bernhard},
	year         = {2011},
	month        = may,
	journal      = {WIREs Computational Molecular Science},
	publisher    = {Wiley},
	volume       = {1},
	number       = {5},
	pages        = {790–809},
	doi          = {10.1002/wcms.34},
	issn         = {1759-0884},
	url          = {http://dx.doi.org/10.1002/wcms.34}
}

@inproceedings{schnet,
	title        = {{SchNet}: A continuous-filter convolutional neural network for modeling quantum interactions},
	author       = {Sch{\"u}tt, Kristof T. and Kindermans, Pieter-Jan and Sauceda, Huziel E. and Chmiela, Stefan and Tkatchenko, Alexandre and M{\"u}ller, Klaus-Robert},
	year         = {2017},
	booktitle    = {Advances in Neural Information Processing Systems},
	publisher    = {Curran Associates, Inc.},
	volume       = {30},
	pages        = {991--1001},
	url          = {https://proceedings.neurips.cc/paper_files/paper/2017/file/303ed4c69846ab36c2904d3ba8573050-Paper.pdf},
	editor       = {Guyon, I. and Von Luxburg, U. and Bengio, S. and Wallach, H. and Fergus, R. and Vishwanathan, S. and Garnett, R.}
}

@article{schneuing2024structure,
	title        = {Structure-based drug design with equivariant diffusion models},
	author       = {Schneuing, Arne and Harris, Charles and Du, Yuanqi and Didi, Kieran and Jamasb, Arian and Igashov, Ilia and Du, Weitao and Gomes, Carla and Blundell, Tom L. and Lio, Pietro and others},
	year         = {2024},
	journal      = {Nature Computational Science},
	publisher    = {Nature Publishing Group US New York},
	volume       = {4},
	number       = {12},
	pages        = {899--909}
}

@article{Schuett2018SchNet,
	title        = {{SchNet -- A} deep learning architecture for molecules and materials},
	author       = {Sch{\"u}tt, Kristof T. and Sauceda, Huziel E. and Kindermans, Pieter-Jan and Tkatchenko, Alexandre and M{\"u}ller, Klaus-Robert},
	year         = {2018},
	journal      = {The Journal of Chemical Physics},
	publisher    = {AIP Publishing},
	volume       = {148},
	number       = {24},
	pages        = {241722},
	doi          = {10.1063/1.5019779}
}

@article{schuett2018schnetpack,
	title        = {{SchNetPack}: A deep learning toolbox for atomistic systems},
	author       = {Sch{\"u}tt, Kristof T. and Kessel, Pan and Gastegger, Michael and Nicoli, Kim A. and Tkatchenko, Alexandre and M{\"u}ller, Klaus-Robert},
	year         = {2018},
	journal      = {Journal of Chemical Theory and Computation},
	publisher    = {American Chemical Society},
	volume       = {15},
	number       = {1},
	pages        = {448--455},
	doi          = {10.1021/acs.jctc.8b00908}
}

@article{schutt2023schnetpack,
	title        = {{SchNetPack 2.0: A neural network toolbox for atomistic machine learning}},
	author       = {Sch{\"u}tt, Kristof T. and Hessmann, Stefaan S. P. and Gebauer, Niklas W. A. and Lederer, Jonas and Gastegger, Michael},
	year         = {2023},
	month        = {04},
	journal      = {The Journal of Chemical Physics},
	volume       = {158},
	number       = {14},
	pages        = {144801},
	doi          = {10.1063/5.0138367},
	issn         = {0021-9606},
	eprint       = {https://pubs.aip.org/aip/jcp/article-pdf/doi/10.1063/5.0138367/16825487/144801_1_5.0138367.pdf}
}

@article{sgdml,
	title        = {sGDML: Constructing accurate and data efficient molecular force fields using machine learning},
	author       = {Chmiela, Stefan and Sauceda, Huziel E. and Poltavsky, Igor and M{\"u}ller, Klaus-Robert and Tkatchenko, Alexandre},
	year         = {2019},
	journal      = {Computer Physics Communications},
	publisher    = {Elsevier},
	volume       = {240},
	pages        = {38--45},
	doi          = {10.1016/j.cpc.2019.02.007}
}

@article{smith2020ani1x_data,
	title        = {The {ANI-1ccx} and {ANI-1x} data sets, coupled-cluster and density functional theory properties for molecules},
	author       = {Smith, Justin S. and Zubatyuk, Roman and Nebgen, Benjamin and Lubbers, Nicholas and Barros, Kipton and Roitberg, Adrian E. and Isayev, Olexandr and Tretiak, Sergei},
	year         = {2020},
	journal      = {Scientific Data},
	publisher    = {Nature Publishing Group UK London},
	volume       = {7},
	number       = {1},
	pages        = {134},
	doi          = {10.1038/s41597-020-0473-z}
}

@inproceedings{so3krates,
	title        = {So3krates: Equivariant attention for interactions on arbitrary length-scales in molecular systems},
	author       = {Frank, J. Thorben and Unke, Oliver T. and M{\"u}ller, Klaus-Robert},
	year         = {2022},
	booktitle    = {Advances in Neural Information Processing Systems},
	publisher    = {Curran Associates, Inc.},
	volume       = {35},
	pages        = {29400--29413},
	url          = {https://proceedings.neurips.cc/paper_files/paper/2022/file/bcf4ca90a8d405201d29dd47d75ac896-Paper-Conference.pdf},
	editor       = {Koyejo, S. and Mohamed, S. and Agarwal, A. and Belgrave, D. and Cho, K. and Oh, A.}
}

@inproceedings{song2019score,
	title        = {Generative Modeling by Estimating Gradients of the Data Distribution},
	author       = {Song, Yang and Ermon, Stefano},
	year         = {2019},
	booktitle    = {Advances in Neural Information Processing Systems},
	publisher    = {Curran Associates, Inc.},
	volume       = {32},
	url          = {https://proceedings.neurips.cc/paper_files/paper/2019/file/3001ef257407d5a371a96dcd947c7d93-Paper.pdf},
	editor       = {Wallach, H. and Larochelle, H. and Beygelzimer, A. and d\textquotesingle Alch{\'e}-Buc, F. and Fox, E. and Garnett, R.}
}

@inproceedings{song2021score,
	title        = {Score-Based Generative Modeling through Stochastic Differential Equations},
	author       = {Song, Yang and Sohl-Dickstein, Jascha and Kingma, Diederik P. and Kumar, Abhishek and Ermon, Stefano and Poole, Ben},
	year         = {2021},
	booktitle    = {International Conference on Learning Representations},
	url          = {https://openreview.net/forum?id=PxTIG12RRHS}
}

@inproceedings{song2023consistency,
	title        = {Consistency models},
	author       = {Song, Yang and Dhariwal, Prafulla and Chen, Mark and Sutskever, Ilya},
	year         = {2023},
	booktitle    = {Proceedings of the 40th International Conference on Machine Learning},
	pages        = {32211--32252}
}

@article{spookynet,
	title        = {Spookynet: Learning force fields with electronic degrees of freedom and nonlocal effects},
	author       = {Unke, Oliver T. and Chmiela, Stefan and Gastegger, Michael and Sch{\"u}tt, Kristof T. and Sauceda, Huziel E. and M{\"u}ller, Klaus-Robert},
	year         = {2021},
	journal      = {Nature Communications},
	publisher    = {Nature Publishing Group UK London},
	volume       = {12},
	number       = {1},
	pages        = {7273},
	doi          = {10.1038/s41467-021-27504-0}
}

@inproceedings{torsional_diff,
	title        = {Torsional Diffusion for Molecular Conformer Generation},
	author       = {Jing, Bowen and Corso, Gabriele and Chang, Jeffrey and Barzilay, Regina and Jaakkola, Tommi S.},
	year         = {2022},
	booktitle    = {Advances in Neural Information Processing Systems},
	url          = {https://openreview.net/forum?id=w6fj2r62r_H},
	editor       = {Oh, Alice H. and Agarwal, Alekh and Belgrave, Danielle and Cho, Kyunghyun}
}

@article{unke2021machine_review,
	title        = {Machine Learning Force Fields},
	author       = {Unke, Oliver T. and Chmiela, Stefan and Sauceda, Huziel E. and Gastegger, Michael and Poltavsky, Igor and Sch{\"u}tt, Kristof T. and Tkatchenko, Alexandre and M{\"u}ller, Klaus-Robert},
	year         = {2021},
	journal      = {Chemical Reviews},
	volume       = {121},
	number       = {16},
	pages        = {10142--10186},
	doi          = {10.1021/acs.chemrev.0c01111}
}

@article{UnkeGems,
	title        = {Biomolecular dynamics with machine-learned quantum-mechanical force fields trained on diverse chemical fragments},
	author       = {Unke, Oliver T. and St{\"o}hr, Martin and Ganscha, Stefan and Unterthiner, Thomas and Maennel, Hartmut and Kashubin, Sergii and Ahlin, Daniel and Gastegger, Michael and Medrano Sandonas, Leonardo and Berryman, Joshua T. and Tkatchenko, Alexandre and M{\"u}ller, Klaus-Robert},
	year         = {2024},
	journal      = {Science Advances},
	volume       = {10},
	number       = {14},
	pages        = {eadn4397},
	doi          = {10.1126/sciadv.adn4397},
	url          = {https://www.science.org/doi/abs/10.1126/sciadv.adn4397},
	eprint       = {https://www.science.org/doi/pdf/10.1126/sciadv.adn4397}
}

@article{von2020exploring,
	title        = {Exploring chemical compound space with quantum-based machine learning},
	author       = {von Lilienfeld, O. Anatole and M{\"u}ller, Klaus-Robert and Tkatchenko, Alexandre},
	year         = {2020},
	journal      = {Nature Reviews Chemistry},
	publisher    = {Nature Publishing Group},
	volume       = {4},
	number       = {7},
	pages        = {347--358},
	doi          = {10.1038/s41570-020-0189-9}
}

@article{wang2025equilibrium,
	title        = {Equilibrium Matching: Generative Modeling with Implicit Energy-Based Models},
	author       = {Wang, Runqian and Du, Yilun},
	year         = {2025},
	journal      = {arXiv preprint arXiv:2510.02300}
}

@article{Watson2023,
	title        = {De novo design of protein structure and function with RFdiffusion},
	author       = {Watson, Joseph L. and Juergens, David and Bennett, Nathaniel R. and Trippe, Brian L. and Yim, Jason and Eisenach, Helen E. and Ahern, Woody and Borst, Andrew J. and Ragotte, Robert J. and Milles, Lukas F. and Wicky, Basile I. M. and Hanikel, Nikita and Pellock, Samuel J. and Courbet, Alexis and Sheffler, William and Wang, Jue and Venkatesh, Preetham and Sappington, Isaac and Torres, Susana V{\'a}zquez and Lauko, Anna and De Bortoli, Valentin and Mathieu, Emile and Ovchinnikov, Sergey and Barzilay, Regina and Jaakkola, Tommi S. and DiMaio, Frank and Baek, Minkyung and Baker, David},
	year         = {2023},
	month        = jul,
	journal      = {Nature},
	publisher    = {Springer Science and Business Media LLC},
	volume       = {620},
	number       = {7976},
	pages        = {1089–1100},
	doi          = {10.1038/s41586-023-06415-8},
	issn         = {1476-4687},
	url          = {http://dx.doi.org/10.1038/s41586-023-06415-8}
}

@article{woo2026riemannian,
	title        = {Riemannian denoising model for molecular structure optimization with chemical accuracy},
	author       = {Woo, Jeheon and Kim, Seonghwan and Kim, Jun Hyeong and Kim, Woo Youn},
	year         = {2026},
	journal      = {Nature Computational Science},
	publisher    = {Nature Publishing Group US New York},
	pages        = {1--11}
}

@inproceedings{zaidi2023pretraining,
	title        = {Pre-training via Denoising for Molecular Property Prediction},
	author       = {Zaidi, Sheheryar and Schaarschmidt, Michael and Martens, James and Kim, Hyunjik and Teh, Yee Whye and Sanchez-Gonzalez, Alvaro and Battaglia, Peter and Pascanu, Razvan and Godwin, Jonathan},
	year         = {2023},
	booktitle    = {The Eleventh International Conference on Learning Representations},
	url          = {https://openreview.net/forum?id=tYIMtogyee}
}

@article{mohan2019robust,
  title={Robust and interpretable blind image denoising via bias-free convolutional neural networks},
  author={Mohan, Sreyas and Kadkhodaie, Zahra and Simoncelli, Eero P. and Fernandez-Granda, Carlos},
  journal={arXiv preprint arXiv:1906.05478},
  year={2019}
}

@article{kadkhodaie2021stochastic,
  title={Stochastic solutions for linear inverse problems using the prior implicit in a denoiser},
  author={Kadkhodaie, Zahra and Simoncelli, Eero},
  journal={Advances in Neural Information Processing Systems},
  volume={34},
  pages={13242--13254},
  year={2021}
}

@article{sun2025noise,
  title={Is noise conditioning necessary for denoising generative models?},
  author={Sun, Qiao and Jiang, Zhicheng and Zhao, Hanhong and He, Kaiming},
  journal={arXiv preprint arXiv:2502.13129},
  year={2025}
}

@article{balcerak2026energy,
  title={Energy matching: unifying flow matching and energy-based models for generative modeling},
  author={Balcerak, Michal and Amiranashvili, Tamaz and Terpin, Antonio and Shit, Suprosanna and Bogensperger, Lea and Kaltenbach, Sebastian and Koumoutsakos, Petros and Menze, Bjoern},
  journal={Advances in Neural Information Processing Systems},
  volume={38},
  pages={8583--8609},
  year={2026}
}

@article{kadkhodaie2026blind,
  title={Blind denoising diffusion models and the blessings of dimensionality},
  author={Kadkhodaie, Zahra and Pooladian, Aram-Alexandre and Chewi, Sinho and Simoncelli, Eero},
  journal={arXiv preprint arXiv:2602.09639},
  year={2026}
}

@article{sahraee2026geometry,
  title={The Geometry of Noise: Why Diffusion Models Don't Need Noise Conditioning},
  author={Sahraee-Ardakan, Mojtaba and Delbracio, Mauricio and Milanfar, Peyman},
  journal={arXiv preprint arXiv:2602.18428},
  year={2026}
}

@misc{ripken2026learning,
  title={Learning Hamiltonian Flow Maps: Mean Flow Consistency for Large-Timestep Molecular Dynamics},      
  author = {Ripken, Winfried and Plainer, Michael and Lied, Gregor and Frank, J. Thorben and Unke, Oliver T. and Chmiela, Stefan and No{\'e}, Frank and M{\"u}ller, Klaus-Robert},
  year         = {2026},
	publisher    = {arXiv},
	url          = {https://arxiv.org/abs/2601.22123},
	copyright    = {arXiv.org perpetual,  non-exclusive license}
}

@inproceedings{bigi2025flashmd,
  author={Bigi, Filippo and Chong, Sanggyu and Kristiadi, Agustinus and Ceriotti, Michele},
 booktitle = {Advances in Neural Information Processing Systems},
 publisher = {Curran Associates, Inc.},
  title={FlashMD: long-stride, universal prediction of molecular dynamics},
 volume = {38},
 year = {2025}
}

@article{thiemann2025forcefree,
    title={Force-Free Molecular Dynamics Through Autoregressive Equivariant Networks},
    author={Thiemann, Fabian L. and Resch{\"u}tzegger, Thiago and Esposito, Massimiliano and Taddese, Tseden and Olarte-Plata, Juan D. and Martelli, Fausto},
    year={2025},
    journal={arXiv:2503.23794},
}

@inproceedings{frans2025onestep,
  title={One-Step Diffusion via Shortcut Models},
  author={Frans, Kevin and Hafner, Danijar and Levine, Sergey and Abbeel, Pieter},
  booktitle={Proceedings of the Thirteenth International Conference on Learning Representations (ICLR)},
  year={2025},
}

@misc{xie2026enhanced,
  title={Enhanced Diffusion Sampling: Efficient Rare Event Sampling and Free Energy Calculation with Diffusion Models}, 	      
  author={Xie, Yu and Winkler, Ludwig and Sun, Lixin and Lewis, Sarah and Foster, Adam E. and Jim{\'e}nez-Luna, Jos{\'e} and Hempel, Tim and Gastegger, Michael and Chen, Yaoyi and Zaporozhets, Iryna and Clementi, Cecilia and Bishop, Christopher M. and No{\'e}, Frank},
	year         = {2026},
	publisher    = {arXiv},
	url          = {https://arxiv.org/abs/2602.16634},
	copyright    = {arXiv.org perpetual,  non-exclusive license}
}

@article{bartok2010gaussian,
  title={Gaussian approximation potentials: The accuracy of quantum mechanics, without the electrons},
  author={Bart{\'o}k, Albert P and Payne, Mike C and Kondor, Risi and Cs{\'a}nyi, G{\'a}bor},
  journal={Physical review letters},
  volume={104},
  number={13},
  pages={136403},
  year={2010},
  publisher={APS}
}

@article{batatia2025foundation,
  title={A foundation model for atomistic materials chemistry},
  author={Batatia, Ilyes and Benner, Philipp and Chiang, Yuan and Elena, Alin M and Kov{\'a}cs, D{\'a}vid P and Riebesell, Janosh and Advincula, Xavier R and Asta, Mark and Avaylon, Matthew and Baldwin, William J and others},
  journal={The Journal of chemical physics},
  volume={163},
  number={18},
  year={2025},
  publisher={AIP Publishing}
}

@article{bartok2017machine,
  title={Machine learning unifies the modeling of materials and molecules},
  author={Bart{\'o}k, Albert P and De, Sandip and Poelking, Carl and Bernstein, Noam and Kermode, James R and Cs{\'a}nyi, G{\'a}bor and Ceriotti, Michele},
  journal={Science advances},
  volume={3},
  number={12},
  pages={e1701816},
  year={2017},
  publisher={American Association for the Advancement of Science}
}

@article{musil2021physics,
  title={Physics-inspired structural representations for molecules and materials},
  author={Musil, Felix and Grisafi, Andrea and Bart{\'o}k, Albert P and Ortner, Christoph and Cs{\'a}nyi, G{\'a}bor and Ceriotti, Michele},
  journal={Chemical reviews},
  volume={121},
  number={16},
  pages={9759--9815},
  year={2021},
  publisher={ACS Publications}
}
\newpage

\appendix
\section{Appendix}

\subsection{Model parameters}
\label{app:hyperparameters}

\begin{table}[htbp]
    \centering
    \caption{Architecture and Training Hyperparameters for GPFF and Baseline Models}
    \label{stab:hyperparameters}
    \begin{tabular}{ll}
        \toprule
        \textbf{Parameter} & \textbf{Value / Description} \\
        \midrule
        Dataset Split & 80\% / 10\% / 10\% (Train / Validation / Test) \\
        Architecture & PaiNN (4 interaction layers) \\
        Feature Dimensions & 256 \\
        Radial Basis Functions & 600 Gaussian RBFs (Cutoff: 150 \AA{}) \\
        Noise Sampling ($\sigma$) & Log-normal: $\ln \sigma \sim \mathcal{N}(-0.7,\, 1.2^2)$, clipped to $[0, 30]$ \\
        Optimizer & AdamW \\
        Learning Rate & $10^{-4}$ (Initial) \\
        Batch Size & 512 \\
        LR Scheduling & Halve LR after 250 epochs without improvement \\
        Early Stopping & 500 epochs without improvement \\
        EMA Decay & 0.995 \\
        SchNetPack Version & 2.1.1 (commit 74940f3) \\
        \bottomrule
    \end{tabular}
\end{table}

\subsection{Sampler parameters}
\label{app:sampler_parameters}

\begin{table}[htbp]
    \centering
    \caption{Sampler hyperparameters used throughout all experiments.}
    \label{stab:hyperparameters_sampling}
    \begin{tabular}{ll}
        \toprule
        \textbf{Parameter} & \textbf{Value} \\
        \midrule
        \multicolumn{2}{l}{\textit{Prior}} \\
        $\sigma_{\max}$ & 30.0~\AA{} \\
        \midrule
        \multicolumn{2}{l}{\textit{EDM noise schedule}} \\
        $\rho$ & 5.0 \\
        $\sigma_{\min}$ & 0.01~\AA{} \\
        $\sigma_{\max}$ & 30.0~\AA{} \\
        \midrule
        \multicolumn{2}{l}{\textit{Stochastic Heun}} \\
        $\sigma_{\mathrm{churn}}$ & 60.0 \\
        $\sigma_{t,\min}$ & 0.01~\AA{} \\
        $\sigma_{t,\max}$ & 15.0~\AA{} \\
        $\sigma_{\mathrm{noise}}$ & 1.0 \\
        \bottomrule
    \end{tabular}
\end{table}

\subsection{Adaptive sampling trajectories}
\label{app:as_trajectories}

\begin{figure}
\centering
\includegraphics[width=1\linewidth]{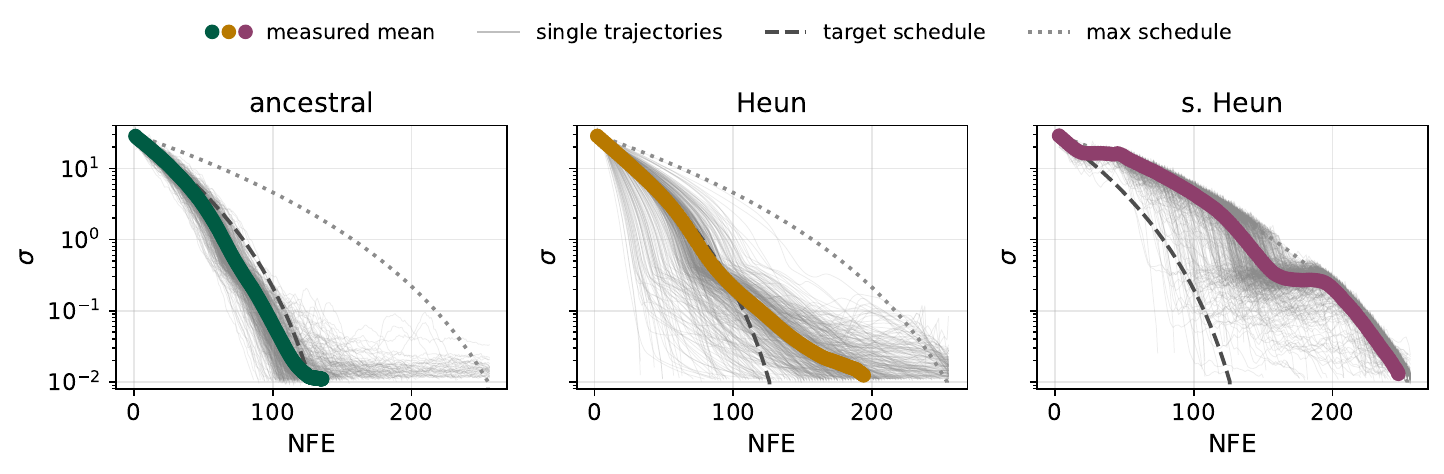}
\caption{\label{sfig:figure_5}Adaptive sampling trajectories for ancestral, Heun, and stochastic Heun sampling with $\Ntarget = \Nsteps / 2$, showing the first 500 generated molecules. Gray lines show individual trajectories of the estimated noise level $\sighat{i}$ over NFE. Colored markers indicate the mean across all trajectories. The dashed line shows the target schedule ($\Ntarget$ steps) and the dotted line the maximum schedule ($\Nsteps$ steps).}
\end{figure}

Figure~\ref{sfig:figure_5} shows adaptive sampling (AS) trajectories of the first 500 generated molecules for ancestral, Heun, and stochastic Heun sampling with $\Ntarget = \Nsteps / 2$.
The individual trajectories (gray lines) illustrate that AS adapts the noise decay to each molecule's denoising difficulty, with faster-converging molecules terminating earlier.
On average (colored markers), the adaptive schedule for ancestral and Heun closely follows the target schedule (dashed line), while the maximum schedule (dotted line) provides the upper bound that prevents stagnation.
Because permutation alignment during training reduces the effective noise level seen by the model, the model's noise-level estimate at inference — where no alignment is applied — tends to fall below the scheduled value, allowing many trajectories to progress faster than the target schedule.
Conversely, near convergence, we observe smaller step sizes and more steps in the final phase.

In contrast to AS with ancestral and deterministic Heun sampling, the AS with stochastic Heun is close to the maximum schedule.
Here, AS mainly provides an advantage in validity of generated molecules at small numbers of target NFEs (see Fig.~\ref{fig:figure_3} in the main text) but converges to a similar amount of NFEs and validity for less ambitious schedules.
We hypothesize that this is caused by the corrective effect of adding noise:
the stochastic Heun sampler increases the noise level to $\tilde{\sigma}^{(i)} = (1+\gamma)\sigi{i}$ before each step.
To avoid an additional inference call, we estimate the pre-injection noise level by rescaling $\sigi{i} = \tilde{\sigma}^{(i)} / (1 + \gamma \alpha)$, where $\alpha$ governs the correction strength.
Due to permutation alignment during training, the model tends to underestimate the added noise, so a correction is necessary; we find that a conservative scaling of $\alpha = 0.5$ provides the best results.

\subsection{Distribution of noise levels}
\label{app:noise_levels}
\begin{figure}[t]
    \centering
    \begin{subfigure}[b]{0.48\textwidth}
        \centering
        \includegraphics[width=\linewidth]{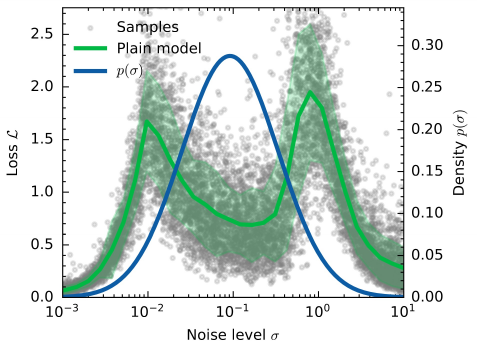}
        \caption{GPFF}
    \end{subfigure}
    \hfill
    \begin{subfigure}[b]{0.48\textwidth}
        \centering
        \includegraphics[width=\linewidth]{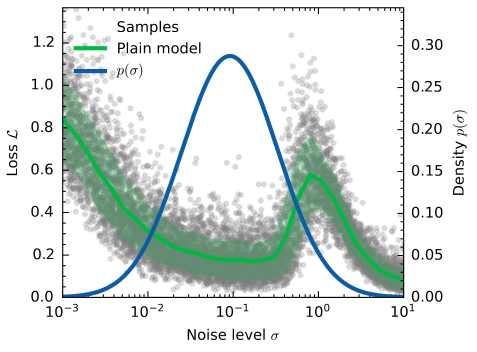}
        \caption{DM}
    \end{subfigure}
    \caption{Loss as a function of noise level $\sigma$ for (a) GPFF and (b) DM, evaluated on preliminary models trained with uniformly sampled $\sigma$. Gray dots show individual test samples, the green line shows the smoothed mean loss, and the blue curve shows the fitted log-normal sampling density $p(\sigma)$ used for the final training runs.}
    \label{fig:app:noise_levels}
\end{figure}

Following Karras~et~al.~\cite{diff_edm}, we concentrate training on informative noise scales by sampling $\sigma \in [0, 30]$ from a log-normal distribution.
To determine the distribution parameters, we first trained preliminary models with uniformly sampled $\sigma$ and evaluated the loss across noise levels on the test set (Figure~\ref{fig:app:noise_levels}).
For GPFF, the loss drops off at small $\sigma$ due to clipping the loss weight $\lambda_{F} = \sigma^{-2}$ to a maximum of 1000 for numerical stability.
From the resulting loss profiles, we chose a log-normal distribution ($\ln \sigma \sim \mathcal{N}(-0.7,\, 1.2^2)$) that concentrates sampling in regions where the loss is highest and the model benefits most from additional training examples.
The calibration was performed independently for GPFF and DM, yielding the same parameters for both.
These parameters were then used to train the final models used in all experiments.

\subsection{Shape predictor} \label{app:cov_predictor}
In direct denoising, standard isotropic Gaussian priors are often insufficient to capture the wide structural variety of molecules.
To address this, we shape the prior with a $3\times3$ covariance matrix $\boldsymbol{\Sigma} \in \mathbb{R}^{3\times3}$ computed from atomic coordinates in their principal frame; the principal variances used in the main text are its diagonal in this frame, $\boldsymbol{\lambda} = \mathrm{diag}(\boldsymbol{\Sigma})$.
This matrix captures the macroscopic molecular geometry, with the relative variance $r_x=\boldsymbol{\lambda}_{x} / \left( \boldsymbol{\lambda}_{x} + \boldsymbol{\lambda}_{y} + \boldsymbol{\lambda}_{z} \right)$ distinguishing between rod, sphere, and disc shapes (see Section \ref{sec:targeted}).

We train a lightweight generator to sample valid covariance matrices conditioned on the number of atoms of a given molecule.
To enforce the required positive definiteness of $\boldsymbol{\Sigma}$, we first parameterize it via the Cholesky decomposition $ \boldsymbol{\Sigma} = \mathbf{L}\mathbf{L}^\top$,
where $\mathbf{L}$ is a lower triangular matrix.
To allow for unconstrained modeling and prediction, we then map $\mathbf{L}$ to a six-dimensional vector $\mathbf{v}$ by applying a logarithmic transformation to its diagonal entries
\begin{equation}
    \mathbf{v} = \left[ \log \left( \mathbf{L}_{11} \right), \mathbf{L}_{21}, \log \left( \mathbf{L}_{22} \right), \mathbf{L}_{31}, \mathbf{L}_{32},
    \log \left( \mathbf{L}_{33} \right)\right]^\top,
\end{equation}
ensuring positive diagonal values and thus positive definiteness of $\boldsymbol{\Sigma}$.

In this unconstrained space of vectors $\mathbf{v}$, we fit a 5-component Gaussian Mixture Model (GMM) \cite{bishopPRML} on the $\mathbf{v}$ extracted from the training data, training a separate GMM for each distinct atom count.
During generation, we first sample an unconstrained vector $\mathbf{v}$ from the GMM corresponding to the target number of atoms, which is then mapped back to the covariance matrix $\boldsymbol{\Sigma}$ and used to draw prior atomic positions $x_i \sim \mathcal{N}(\mathbf{0}, \boldsymbol{\Sigma})$.

This lightweight shape predictor trains in seconds and introduces negligible computational overhead to the overall generative pipeline.

\subsection{Principal moments of inertia} \label{app:pmi}
Each shape target in Section~\ref{sec:targeted} is a triple of the relative variances $(r_x, r_y, r_z)$ defined in Section~\ref{app:cov_predictor}, evaluated in the principal frame of $\boldsymbol{\Sigma}_{\mathrm{target}}$ so that the components sum to one and are sorted in non-increasing order.
We pick one canonical target per archetype: rod $=(0.9,\, 0.05,\, 0.05)$, sphere $=(1/3,\, 1/3,\, 1/3)$, and disc $=(0.5,\, 0.5,\, 0)$.
The rod target is offset from the vertex $(1, 0, 0)$, since concentrating all variance on a single axis is not feasible for most molecules.

For Figure~\ref{fig:covariance_generation} we follow Sauer and Schwarz~\cite{sauer2003molecular} and compute the principal moments of inertia (PMI) $I_1 \leq I_2 \leq I_3$ with uniform unit weights rather than atomic masses, so the descriptor reflects atomic geometry alone and is consistent with how $\boldsymbol{\Sigma}_{\mathrm{target}}$ is defined.
In the principal frame of each molecule's empirical covariance $\boldsymbol{\Sigma}$ with centered coordinates, the moments about the longest, middle, and shortest axis simplify to $I_1 \propto \boldsymbol{\Sigma}_{yy} + \boldsymbol{\Sigma}_{zz}$, $I_2 \propto \boldsymbol{\Sigma}_{xx} + \boldsymbol{\Sigma}_{zz}$, and $I_3 \propto \boldsymbol{\Sigma}_{xx} + \boldsymbol{\Sigma}_{yy}$ (with $\boldsymbol{\Sigma}_{xx} \geq \boldsymbol{\Sigma}_{yy} \geq \boldsymbol{\Sigma}_{zz}$), so the normalized PMI ratios (NPRs) become
\begin{equation}
\mathrm{NPR}_1 = \frac{I_1}{I_3} = \frac{\boldsymbol{\Sigma}_{yy} + \boldsymbol{\Sigma}_{zz}}{\boldsymbol{\Sigma}_{xx} + \boldsymbol{\Sigma}_{yy}}, \qquad \mathrm{NPR}_2 = \frac{I_2}{I_3} = \frac{\boldsymbol{\Sigma}_{xx} + \boldsymbol{\Sigma}_{zz}}{\boldsymbol{\Sigma}_{xx} + \boldsymbol{\Sigma}_{yy}}.
\end{equation}
Each molecule maps to a point $(\mathrm{NPR}_1, \mathrm{NPR}_2)$ in the rod--sphere--disc triangle with vertices $\text{Rod}=(0,1)$, $\text{Sphere}=(1,1)$, $\text{Disc}=(0.5,0.5)$, to which the sphere and disc targets map exactly while the offset rod target lands at $(0.105, 1)$.

\subsection{Training of property models} \label{app:property_models}
We trained separate PaiNN property predictors on QM9 for the HOMO--LUMO gap and $U_0$ energy used in the JS divergence evaluation.
Both models use 128 hidden features, 3 interaction blocks, a 5.0~\AA{} cutoff with 20 Gaussian radial basis functions, and a cosine cutoff function.
Training was performed with AdamW at a learning rate of $5 \times 10^{-4}$ and batch size 100, with the learning rate halved after 75 epochs without improvement and early stopping after 200 epochs without improvement, using EMA with decay 0.995 and an 80/10/10 train/validation/test split.
The models achieve test MAEs of 42.4~meV for the gap and 6.2~meV for $U_0$.
For the $U_0$ distribution comparison, we subtract atomic reference energies (atomrefs) from the predictions, so that the distribution reflects only bonding and geometry rather than composition (number and types of atoms), which is an input to the generation rather than an output.

\subsection{Permutation alignment}
\label{app:permutation_alignment}
During training, for each reference geometry $\mathbf{X}$ a perturbed geometry $\X$ is sampled from the forward process.
For GPFF, we align the atom ordering of $\X$ to $\mathbf{X}$ before computing the pseudo-forces.
The alignment is performed independently per atom type using the Hungarian algorithm on pairwise squared distances between centered coordinates.

\begin{figure}
\centering
\includegraphics[width=1\linewidth]{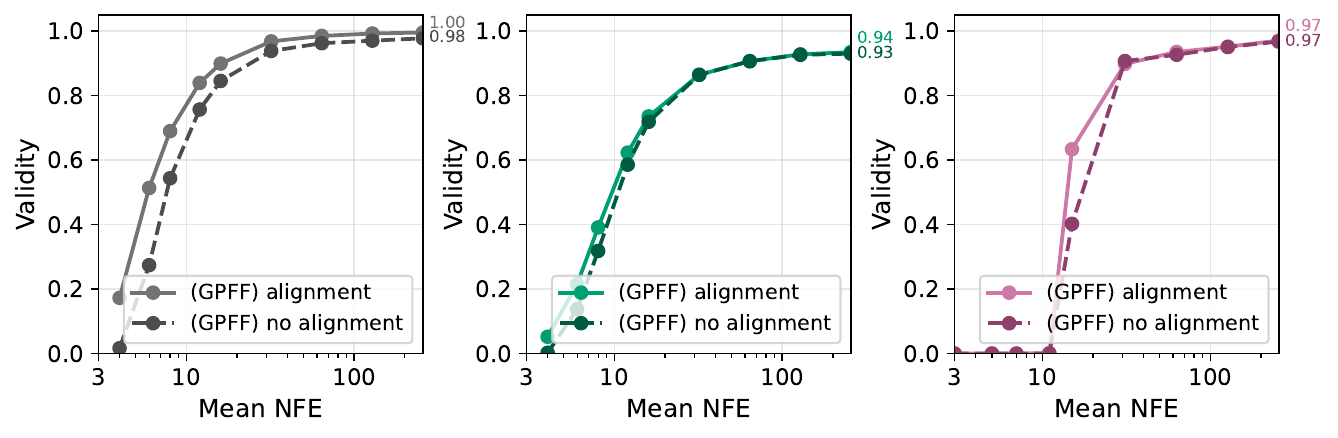}
\caption{\label{sfig:figure_1}Effect of atom index alignment during training on GPFF sampling. Validity as a function of NFE for direct denoising (left), ancestral sampling (middle), and stochastic Heun sampling (right), comparing alignment and no alignment.}
\end{figure}

Figure~\ref{sfig:figure_1} compares GPFF trained with and without permutation alignment across three samplers: DD, ancestral, and stochastic Heun.
The aligned model consistently achieves slightly higher validity, with the advantage most visible at low NFE.
At convergence, the differences are small (1.00 vs.\ 0.98 for DD, 0.94 vs.\ 0.93 for ancestral, 0.97 vs.\ 0.97 for stochastic Heun), indicating that alignment provides a modest but consistent improvement.

\subsection{Ablation studies on time step for GPFF}
\label{app:ablation_t}
\begin{figure}
\centering
\includegraphics[width=1\linewidth]{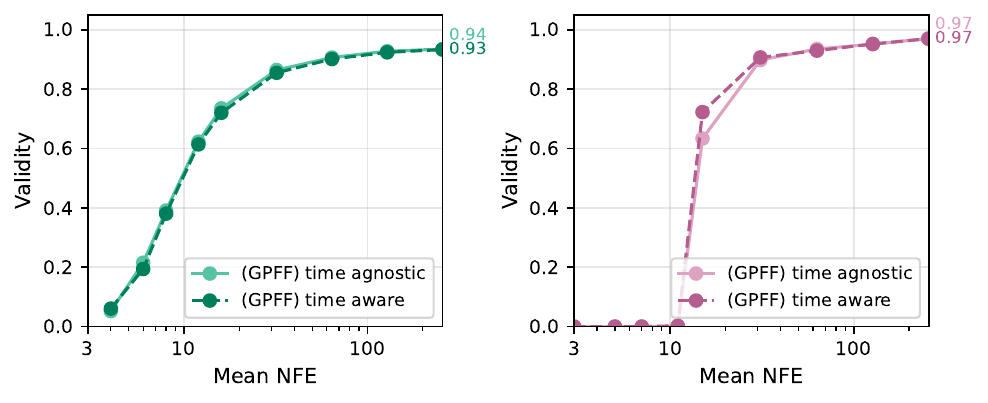}
\caption{\label{sfig:figure_2}Impact of time-step conditioning of GPFF models on sampling quality. Validity as a function of NFE for ancestral sampling (left) and stochastic Heun sampling (right), comparing time-agnostic and time-aware GPFF models.}
\end{figure}

We evaluate whether explicit time conditioning improves GPFF sampling quality by training a time-aware GPFF variant with identical hyperparameters, differing only in the use of the same $t$-conditioned output head as the DM.
Figure~\ref{sfig:figure_2} shows validity across matched NFE sweeps for ancestral and stochastic Heun sampling.
The results show no meaningful difference between the two variants (0.94 vs.\ 0.93 for ancestral, 0.97 vs.\ 0.97 for stochastic Heun), confirming that explicit time conditioning is unnecessary for GPFF and that the model captures sufficient information about the noise level from the geometry alone.

\subsection{Ablation studies on time step for diffusion models}
\label{app:dm_time_ablation}
\begin{figure}
\centering
\includegraphics[width=1\linewidth]{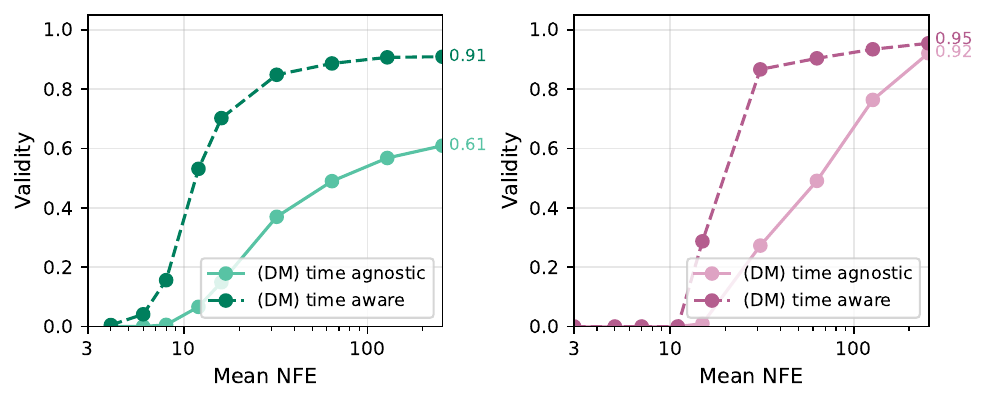}
\caption{\label{sfig:figure_3}Impact of time-step conditioning of DMs on sampling quality. Validity as a function of NFE for ancestral sampling (left) and stochastic Heun sampling (right), comparing time-agnostic and time-aware DMs.}
\end{figure}

We perform the converse experiment for the DM by training a time-agnostic variant with identical hyperparameters, replacing the $t$-conditioned output head with the gated equivariant head used by GPFF.
As shown in Figure~\ref{sfig:figure_3}, the time-agnostic DM performs substantially worse than its time-aware counterpart, reaching only 0.61 vs.\ 0.91 validity for ancestral sampling and 0.92 vs.\ 0.95 for stochastic Heun.
While stochastic Heun partially recovers at higher NFE, the gap remains large for ancestral sampling across the entire NFE range.
We attribute this asymmetry to a fundamental difference in what the two model types learn.
GPFF predicts a quantity whose magnitude scales with the noise level, effectively forcing the model to implicitly encode the time step in its predictions, as demonstrated by the noise-level estimator used in AS (Equation~\ref{noise_estimate}).
The DM, by contrast, predicts a normalized direction that is magnitude-free and therefore has no incentive to learn the current noise level from the geometry alone.
Without explicit $t$-conditioning, the DM lacks the information needed to produce appropriately scaled score estimates, leading to degraded sampling quality.

\subsection{Examples of property distributions}\label{app:histograms}
\begin{figure}
\centering
\includegraphics[width=1\linewidth]{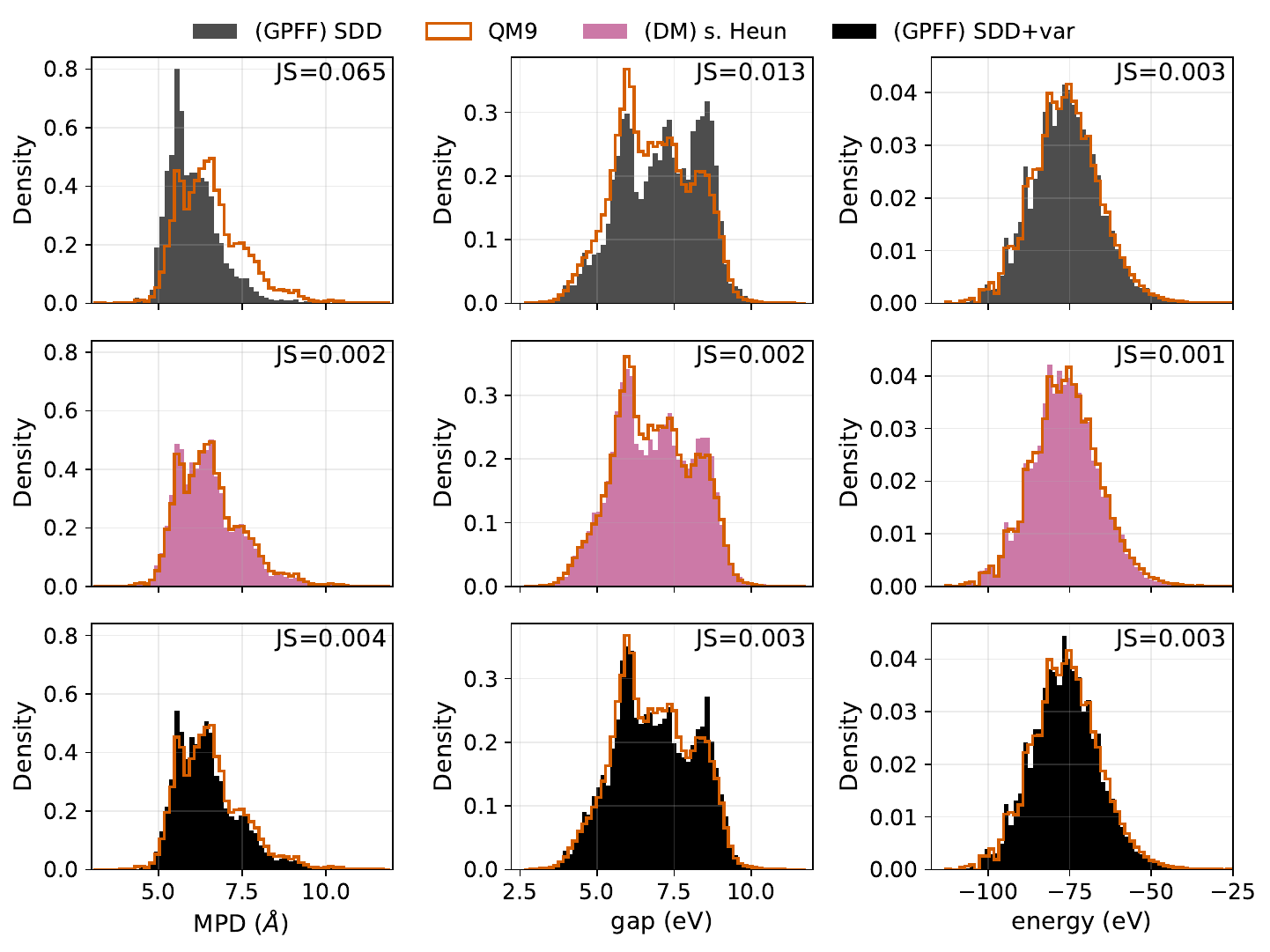}
\caption{\label{sfig:figure_4}Distribution comparison to QM9 at 256 NFE. Rows correspond to direct denoising (GPFF), stochastic Heun (DM), and direct denoising with covariance enforcement (GPFF). Columns show the distributions of MPD, HOMO-LUMO gap, and energy U0. Orange histograms show generated valid molecules and black outlines show the QM9 target distribution; the JS divergence for each panel is reported in the top-right corner.}
\end{figure}

Figure~\ref{sfig:figure_4} compares the property distributions of valid generated molecules against QM9 at 256 NFE for three samplers: GPFF direct denoising (DD, from Section~\ref{subsec:direct_denoising}), the DM stochastic Heun baseline (from Section~\ref{subsec:direct_denoising}), and GPFF with covariance-enforced direct denoising (DD+shape, from Section~\ref{subsec:results_shape_bias}).
The MPD bias of standard DD is clearly visible, with a JS divergence of 0.065 compared to 0.002 for stochastic Heun and 0.004 for DD+shape.
The HOMO--LUMO gap distribution is also affected in the DD variant (JS = 0.013 vs.\ 0.002 and 0.003), which we attribute to the gap's dependence on molecular shape.
The atomization energy $U_0$, by contrast, is largely unaffected by the MPD bias, showing comparable JS divergences across all three methods (0.003, 0.001, 0.003).
Although stochastic Heun achieves marginally lower JS divergence than DD+shape, the differences are negligible at these levels and not visible in the histograms.

\subsection{Pseudocode for sampling algorithms} \label{app:sampling_algorithms}

\begin{algorithm}
\caption{Direct Denoising}
\begin{algorithmic}[1]
\Procedure{DirectDenoising}{$\Xrev{0}, \text{GPFF}, \Nsteps, f_{\max}, \text{enforce\_shape}, \text{stochastic}, \boldsymbol{\lambda}_{\mathrm{target}}, p$}
    \For{$i = 0$ \textbf{to} $\Nsteps-1$}
        \If{enforce\_shape} \Comment{Apply shape constraint}
            \State $\alpha \gets (i / \Nsteps)^p$
            \State $\boldsymbol{\tilde{\lambda}} \gets \text{PrincipalVariances}(\Xrev{i})$
            \State $\Xrev{i} \gets (1-\alpha) \cdot \sqrt{\boldsymbol{\lambda}_{\mathrm{target}} / \boldsymbol{\tilde{\lambda}}} \cdot \Xrev{i} + \alpha \cdot \Xrev{i}$        \EndIf
        \State $\{\Xpred, \Fpred\} \gets \text{GPFF}(\Xrev{i})$ \Comment{Predict clean geometry and forces}
        \State $\Xrev{i+1} \gets \Xpred$
        \If{$\|\Fpred\|_{\max} \leq f_{\max}$} \Comment{Force-based stopping criterion}
            \State \textbf{break}
        \EndIf
        \If{\text{stochastic}}
            \State $\beta \gets 1 - i / \Nsteps$ \Comment{Linearly decaying noise scale}
            \State $\Xrev{i+1} \gets \Xrev{i+1} + \beta \cdot \eps$, \quad $\eps \sim \mathcal{N}(\mathbf{0}, \mathbf{I})$ \Comment{Stochastic noise injection}
        \EndIf
    \EndFor
    \State \Return $\Xrev{i+1}$
\EndProcedure
\end{algorithmic}
\end{algorithm}

\begin{algorithm}
\caption{Ancestral Sampler}
\begin{algorithmic}[1]
\Procedure{Ancestral}{$\Xrev{0}, \text{GPFF}, \{\sigi{i}\}_{i=0}^{\Nsteps}$}
    \For{$i = 0$ \textbf{to} $\Nsteps-1$}
        \State $\Fpred \gets \text{GPFF}(\Xrev{i})$ \Comment{Predict pseudo-forces}
        \State $\spred \gets \Fpred / (-2 \sigi{i})$ \Comment{Convert pseudo-forces to score}
        \State $\Xrev{i+1} \gets \Xrev{i} + \spred \cdot ((\sigi{i})^2 - (\sigi{i+1})^2)$ \Comment{Deterministic score update}
        \If{$\sigi{i+1} > 0$}
            \State $\sigma_{\mathrm{inj}} \gets \sqrt{\frac{(\sigi{i+1})^2 ((\sigi{i})^2 - (\sigi{i+1})^2)}{(\sigi{i})^2}}$ \Comment{Noise injection scale}
            \State $\Xrev{i+1} \gets \Xrev{i+1} + \sigma_{\mathrm{inj}} \cdot \eps$, \quad $\eps \sim \mathcal{N}(\mathbf{0}, \mathbf{I})$ \Comment{Add corrective noise}
        \EndIf
    \EndFor
    \State \Return $\Xrev{i+1}$
\EndProcedure
\end{algorithmic}
\end{algorithm}

\begin{algorithm}
\caption{Adaptive Ancestral Sampler}
\begin{algorithmic}[1]
\Procedure{AdaptiveAncestral}{$\Xrev{0}, \text{GPFF}, \Ntarget, \Nsteps$}
    \State $\Delta s \gets \frac{1}{\Ntarget - 1}$ \Comment{Initial step size from target schedule}
    \For{$i = 0$ \textbf{to} $\Nsteps-1$}
        \State \textit{// 1. Estimate noise level and adapt schedule}
        \State $\{\Fpred, \sighat{i}\} \gets \text{GPFF}(\Xrev{i})$ \Comment{Predict pseudo-forces and estimate $\sigma$}
        \If{$i > 0$}
            \State $\Delta s \gets \frac{(\sighat{i-1})^{1/\rho} - (\sighat{i})^{1/\rho}}{\sigma_{\max}^{1/\rho} - \sigma_{\min}^{1/\rho}}$ \Comment{Observed step size}
        \EndIf
        \State $\sigi{i+1} \gets \left((\sighat{i})^{1/\rho} - \Delta s \cdot (\sigma_{\max}^{1/\rho} - \sigma_{\min}^{1/\rho})\right)^{\rho}$
        \State $\sigi{i+1} \gets \min\left(\sigi{i+1},\; \sigma_{\mathrm{upper}}^{(i+1)}\right)$ \Comment{Bound by maximum schedule}
        \If{$\sigi{i+1} \leq \sigma_{\mathrm{min}}$}
            \State $\sigi{i+1} = 0$
        \EndIf
        \State
        \State \textit{// 2. Score update}
        \State $\spred \gets \Fpred / (-2 \sighat{i})$ \Comment{Convert pseudo-forces to score}
        \State $\Xrev{i+1} \gets \Xrev{i} + \spred \cdot ((\sighat{i})^2 - (\sigi{i+1})^2)$ \Comment{Deterministic score update}
        \If{$\sigi{i+1} > 0$}
            \State $\sigma_{\mathrm{inj}} \gets \sqrt{\frac{(\sigi{i+1})^2 ((\sighat{i})^2 - (\sigi{i+1})^2)}{(\sighat{i})^2}}$ \Comment{Noise injection scale}
            \State $\Xrev{i+1} \gets \Xrev{i+1} + \sigma_{\mathrm{inj}} \cdot \eps$, \quad $\eps \sim \mathcal{N}(\mathbf{0}, \mathbf{I})$ \Comment{Add corrective noise}
        \EndIf
    \EndFor
    \State \Return $\Xrev{i+1}$
\EndProcedure
\end{algorithmic}
\end{algorithm}

\begin{algorithm}
\caption{Deterministic Heun Sampler}
\begin{algorithmic}[1]
\Procedure{HeunSampler}{$\Xrev{0}, \text{GPFF}, \{\sigi{i}\}_{i=0}^{\Nsteps}$}
    \For{$i = 0$ \textbf{to} $\Nsteps-1$}
        \State $\Xpred' \gets \text{GPFF}(\Xrev{i})$ \Comment{Predict clean geometry}
        \State $\mathbf{d}' \gets \frac{1}{\sigi{i}} \left( \Xrev{i} - \Xpred' \right)$ \Comment{Evaluate $d\mathbf{X}/d\sigma$ at $\sigi{i}$}
        \State $\Xrev{i+1} \gets \Xrev{i} + (\sigi{i+1} - \sigi{i}) \mathbf{d}'$ \Comment{Euler predictor step}
        
        \If{$i == N-1$} \Comment{Convergence criterion}
            \State \textbf{break}
        \Else
        \State $\Xpred'' \gets \text{GPFF}(\Xrev{i+1})$ \Comment{Predict at new position}
        \State $\mathbf{d}'' \gets \frac{1}{\sigi{i+1}} \left( \Xrev{i+1} - \Xpred'' \right)$
        \State $\Xrev{i+1} \gets \Xrev{i} + (\sigi{i+1} - \sigi{i}) \left( \frac{1}{2}\mathbf{d}' + \frac{1}{2}\mathbf{d}'' \right)$ \Comment{Corrector step}
        \EndIf
    \EndFor
    \State \Return $\Xrev{i+1}$
\EndProcedure
\end{algorithmic}
\end{algorithm}

\begin{algorithm}
\caption{Adaptive Heun Sampler}
\begin{algorithmic}[1]
\Procedure{AdaptiveHeun}{$\Xrev{0}, \text{GPFF}, \{\sigi{i}_{\mathrm{upper}}\}_{i=0}^{\Nsteps}, \Ntarget, \sigma_{\min}, \sigma_{\max}, \rho$}
    \State $\Delta s \gets \frac{1}{\Ntarget - 1}$ \Comment{Initial step size from target schedule}
    \For{$i = 0$ \textbf{to} $\Nsteps-1$}
        \State $\{\Xpred', \sighat{i}\} \gets \text{GPFF}(\Xrev{i})$ \Comment{Predict clean geometry and estimate noise level}
        \If{$i > 0$} \Comment{Estimate step size}
            \State $\Delta s \gets \frac{{(\sighat{i-1}})^{1/\rho} - {(\sighat{i}})^{1/\rho}}{\sigma_{\max}^{1/\rho} - \sigma_{\min}^{1/\rho}}$
        \EndIf
        \State $\sigi{i+1} \gets \min\left(({(\sighat{i}})^{1/\rho} + \Delta s \cdot (\sigma_{\min}^{1/\rho} - \sigma_{\max}^{1/\rho}))^\rho,\; \sigma_{\mathrm{upper}}^{(i+1)}\right)$ \Comment{Compute next noise level}
        \If{$\sigi{i+1} \leq \sigma_{\mathrm{min}}$}
            \State $\sigi{i+1} = 0$
        \EndIf
        \State $\mathbf{d}' \gets \frac{1}{\sighat{i}} \left( \Xrev{i} - \Xpred' \right)$
        \State $\Xrev{i+1} \gets \Xrev{i} + (\sigi{i+1} - \sighat{i}) \mathbf{d}'$ \Comment{Step 1: Update positions}
        \If{$\sighat{i} \leq \sigma_{\mathrm{min}}$} \Comment{Check convergence}
            \State \textbf{break}
        \Else
        \State $\Xpred'' \gets \text{GPFF}(\Xrev{i+1})$ \Comment{Predict clean geometry}
        \State $\mathbf{d}'' \gets \frac{1}{\sigi{i+1}} \left( \Xrev{i+1} - \Xpred'' \right)$
        \State $\Xrev{i+1} \gets \Xrev{i} + (\sigi{i+1} - \sighat{i}) \left( \frac{1}{2}\mathbf{d}' + \frac{1}{2}\mathbf{d}'' \right)$ \Comment{Step 2: Corrector}
        \EndIf

    \EndFor
    \State \Return $\Xrev{i+1}$
\EndProcedure
\end{algorithmic}
\end{algorithm}

\begin{algorithm}
\caption{Stochastic Heun Sampler}
\begin{algorithmic}[1]
\Procedure{StochasticHeun}{$\Xrev{0}, \text{GPFF}, \{\sigi{i}\}_{i=0}^{\Nsteps}, \sigma_{\mathrm{churn}}, \sigma_{t,\min}, \sigma_{t,\max}, \sigma_{\mathrm{noise}}$}
    \For{$i = 0$ \textbf{to} $\Nsteps-1$}
        \State $\gamma^{(i)} \gets 0$
        \If{$\sigma_{t,\min} \leq \sigi{i} \leq \sigma_{t,\max}$}
            \State $\gamma^{(i)} \gets \min\left(\sigma_{\mathrm{churn}} / \Nsteps,\; \sqrt{2}-1\right)$
        \EndIf
        \State $\tilde{\sigma}^{(i)} \gets \sigi{i} (1 + \gamma^{(i)})$ \Comment{Increase noise level}
        \State $\Xrev{i} \gets \Xrev{i} + \sigma_{\mathrm{noise}} \cdot \eps \sqrt{(\tilde{\sigma}^{(i)})^2 - (\sigi{i})^2}$, \quad $\eps \sim \mathcal{N}(\mathbf{0}, \mathbf{I})$ \Comment{Inject noise}
        \State $\Xpred' \gets \text{GPFF}(\Xrev{i})$ \Comment{Predict clean geometry}
        \State $\mathbf{d}' \gets \frac{1}{\hat{\sigma}} (\Xrev{i} - \Xpred')$ \Comment{Evaluate $d\mathbf{X}/d\sigma$ at $\tilde{\sigma}^{(i)}$}
        \State $\Xrev{i+1} \gets \Xrev{i} + (\sigi{i+1} - \hat{\sigma}) \mathbf{d}'$ \Comment{Euler predictor step}
        \If{$\sigi{i+1} == 0$}
            \State \textbf{break}
        \Else
            \State $\Xpred'' \gets \text{GPFF}(\Xrev{i+1})$ \Comment{Predict at new position}
            \State $\mathbf{d}'' \gets \frac{1}{\sigi{i+1}} (\Xrev{i+1} - \Xpred'')$
            \State $\Xrev{i+1} \gets \Xrev{i+1} + (\sigi{i+1} - \hat{\sigma}) \left( \frac{1}{2}\mathbf{d}' + \frac{1}{2}\mathbf{d}'' \right)$ \Comment{Corrector step}
        \EndIf
    \EndFor
    \State \Return $\Xrev{i+1}$
\EndProcedure
\end{algorithmic}
\end{algorithm}

\begin{algorithm}
\caption{Adaptive Stochastic Heun Sampler}
\begin{algorithmic}[1]
\Procedure{AdaptiveStochasticHeun}{$\Xrev{0}, \text{GPFF}, \Ntarget, \Nsteps, \sigma_{\mathrm{churn}}, \sigma_{t,\min}, \sigma_{t,\max}, \sigma_{\mathrm{noise}}, \alpha$}
    \State $\Delta s \gets \frac{1}{\Ntarget - 1}$ \Comment{Initial step size from target schedule}
    \For{$i = 0$ \textbf{to} $\Nsteps-1$}
        \State \textit{// 1. Estimate noise level and adapt schedule}
        \State $\{\Xpred, \sighat{i}\} \gets \text{GPFF}(\Xrev{i})$ \Comment{Predict clean geometry and estimate $\sigma$}
        \If{$i > 0$}
            \State $\Delta s \gets \frac{(\sighat{i-1})^{1/\rho} - (\sighat{i})^{1/\rho}}{\sigma_{\max}^{1/\rho} - \sigma_{\min}^{1/\rho}}$ \Comment{Observed step size}
        \EndIf
        \State $\sigi{i+1} \gets \left((\sighat{i})^{1/\rho} - \Delta s \cdot (\sigma_{\max}^{1/\rho} - \sigma_{\min}^{1/\rho})\right)^{\rho}$
        \State $\sigi{i+1} \gets \min\left(\sigi{i+1},\; \sigma_{\mathrm{upper}}^{(i+1)}\right)$ \Comment{Bound by maximum schedule}
        
        \State \textit{// 2. Stochastic noise injection}
        \State $\gamma^{(i)} \gets 0$
        \If{$\sigma_{t,\min} \leq \sighat{i} \leq \sigma_{t,\max}$}
            \State $\gamma^{(i)} \gets \min\left(\sigma_{\mathrm{churn}} / \Nsteps,\; \sqrt{2}-1\right)$
        \EndIf
        \State $\tilde{\sigma}^{(i)} \gets \sighat{i} (1 + \gamma^{(i)})$ \Comment{Increase noise level}
        \State $\Xrev{i} \gets \Xrev{i} + \sigma_{\mathrm{noise}} \cdot \eps \sqrt{(\tilde{\sigma}^{(i)})^2 - (\sighat{i})^2}$, \quad $\eps \sim \mathcal{N}(\mathbf{0}, \mathbf{I})$ \Comment{Inject noise}
        \State
        \State \textit{// 3. Correct noise estimate for injected noise}
        \State $\sighat{i} \gets \tilde{\sigma}^{(i)} / (1 + \gamma^{(i)} \cdot \alpha)$ \Comment{Rescale with correction factor $\alpha$}
        \State
        \State \textit{// 4. Euler predictor step}
        \State $\mathbf{d}' \gets \frac{1}{\tilde{\sigma}^{(i)}} (\Xrev{i} - \Xpred)$
        \State $\Xrev{i+1} \gets \Xrev{i} + (\sigi{i+1} - \tilde{\sigma}^{(i)}) \mathbf{d}'$
        \If{$\sigi{i+1} == 0$}
            \State \textbf{break}
        \Else
            \State \textit{// 5. Corrector step}
            \State $\Xpred'' \gets \text{GPFF}(\Xrev{i+1})$ \Comment{Predict at new position}
            \State $\mathbf{d}'' \gets \frac{1}{\sigi{i+1}} (\Xrev{i+1} - \Xpred'')$
            \State $\Xrev{i+1} \gets \Xrev{i} + (\sigi{i+1} - \tilde{\sigma}^{(i)}) \left( \frac{1}{2}\mathbf{d}' + \frac{1}{2}\mathbf{d}'' \right)$ \Comment{Corrector step}
        \EndIf
    \EndFor
    \State \Return $\Xrev{i+1}$
\EndProcedure
\end{algorithmic}
\end{algorithm}

\end{document}